\documentclass[10pt,twocolumn,letterpaper]{article}

\usepackage{iccv}

\usepackage{amsmath,amsfonts,bm}

\def\eqref#1{equation~\ref{#1}}

\def\1{\bm{1}}

\def\rvx{{\mathbf{x}}}

\DeclareMathAlphabet{\mathsfit}{\encodingdefault}{\sfdefault}{m}{sl}
\SetMathAlphabet{\mathsfit}{bold}{\encodingdefault}{\sfdefault}{bx}{n}

\def\gC{{\mathcal{C}}}

\def\gL{{\mathcal{L}}}

\def\gR{{\mathcal{R}}}

\def\gY{{\mathcal{Y}}}

\DeclareMathOperator*{\argmin}{arg\,min}

\usepackage{times}
\usepackage{epsfig}
\usepackage{graphicx}
\usepackage{amsmath}
\usepackage{amssymb}
\usepackage{booktabs}
\usepackage{acro}
\usepackage{amsfonts}
\usepackage{algorithm2e}
\usepackage{multirow}
\usepackage{booktabs}
\usepackage[table,xcdraw]{xcolor}
\usepackage[pagebackref=true,breaklinks=true,letterpaper=true,colorlinks,bookmarks=false]{hyperref}

\usepackage[capitalize]{cleveref}
\crefname{section}{Sec.}{Secs.}
\Crefname{section}{Section}{Sections}
\Crefname{table}{Table}{Tables}
\crefname{table}{Tab.}{Tabs.}

\SetKwInput{KwData}{Input}
\SetKwInput{KwResult}{Output}

\usepackage{xr}
\DeclareAcronym{ge}{
short=GE,
long=General Encoder,
foreign-plural={}
}

\DeclareAcronym{gs}{
short=ContinualSeg,
long=Continual Segment,
foreign-plural={}
}

\DeclareAcronym{fov}{
short=FOV,
long=field of view,
foreign-plural={}
}

\DeclareAcronym{fbkg}{
short=FBkg,
long=fake background,
foreign-plural={}
}

\DeclareAcronym{kd}{
short=KD,
long=knowledge distillation,
foreign-plural={}
}

\def\highlight{black} 
\def\edited{black} 

\iccvfinalcopy 

\def\iccvPaperID{5269} 

\ificcvfinal\fi

\makeatletter
\def\thanks#1{\protected@xdef\@thanks{\@thanks\protect\footnotetext{#1}}}
\makeatother

\begin{document}

\title{Continual Segment: Towards a Single, Unified and Non-forgetting Continual Segmentation Model of 143 Whole-body Organs in CT Scans}

\author{Zhanghexuan Ji$^{1,2\dagger}$\,\, Dazhou Guo$^{1\dagger}$ \,\, Puyang Wang$^{1}$ \,\, Ke Yan$^{1,3}$ \,\, Le Lu$^{1}$ \,\, Minfeng Xu$^{1,3}$ \\ Jingren Zhou$^{1,3}$ \,\, Qifeng Wang$^{4}$\,\, Jia Ge$^{5}$ \,\,  Mingchen Gao$^{2}$ \,\, Xianghua Ye$^{5*}$ \,\,  Dakai Jin$^{1*}$\\ 
\vspace{-3mm}
\\  
$^1$DAMO Academy, Alibaba Group \,\, $^2$ University at Buffalo \,\, $^3$Hupan Lab, 310023, Hangzhou, China \\  $^4$Sichuan Cancer Hospital \,\, $^5$The First Affiliated Hospital of Zhejiang University
\thanks{$\dagger$ ZJ and DG contribute equally. $*$ For correspondence, please contact XY (hye1982@zju.edu.cn) and DJ (dakai.jin@alibaba-inc.com). \\ \textit{This paper is accepted by ICCV 2023}}
}

\maketitle

\begin{abstract}

Deep learning empowers the mainstream medical image segmentation methods. Nevertheless, current deep segmentation approaches are not capable of efficiently and effectively adapting and updating the trained models when new segmentation classes are incrementally added. In the real clinical environment, it can be preferred that segmentation models could be dynamically extended to segment new organs/tumors without the (re-)access to previous training datasets due to obstacles of patient privacy and data storage. This process can be viewed as a continual semantic segmentation (CSS) problem, being understudied for multi-organ segmentation. In this work, we propose a new architectural CSS learning framework to learn a single deep segmentation model for segmenting a total of 143 whole-body organs. Using the encoder/decoder network structure, we demonstrate that a continually trained then frozen encoder coupled with incrementally-added decoders can extract sufficiently representative image features for new classes to be subsequently and validly segmented, while avoiding the catastrophic forgetting in CSS. To maintain a single network model complexity, each decoder is progressively pruned using neural architecture search and teacher-student based knowledge distillation. \textcolor{\edited}{Finally, we propose a body-part and anomaly-aware output merging module to combine organ predictions originating from different decoders and incorporate both healthy and pathological organs appearing in different datasets.} Trained and validated on 3D CT scans of 2500+ patients from four datasets, our single network can segment a total of 143 whole-body organs  with very high accuracy, closely reaching the upper bound performance level by training four separate segmentation models (i.e., one model per dataset/task).

\end{abstract}

    \begin{figure}[ht]
    \centering
    \includegraphics[width=0.95\columnwidth]{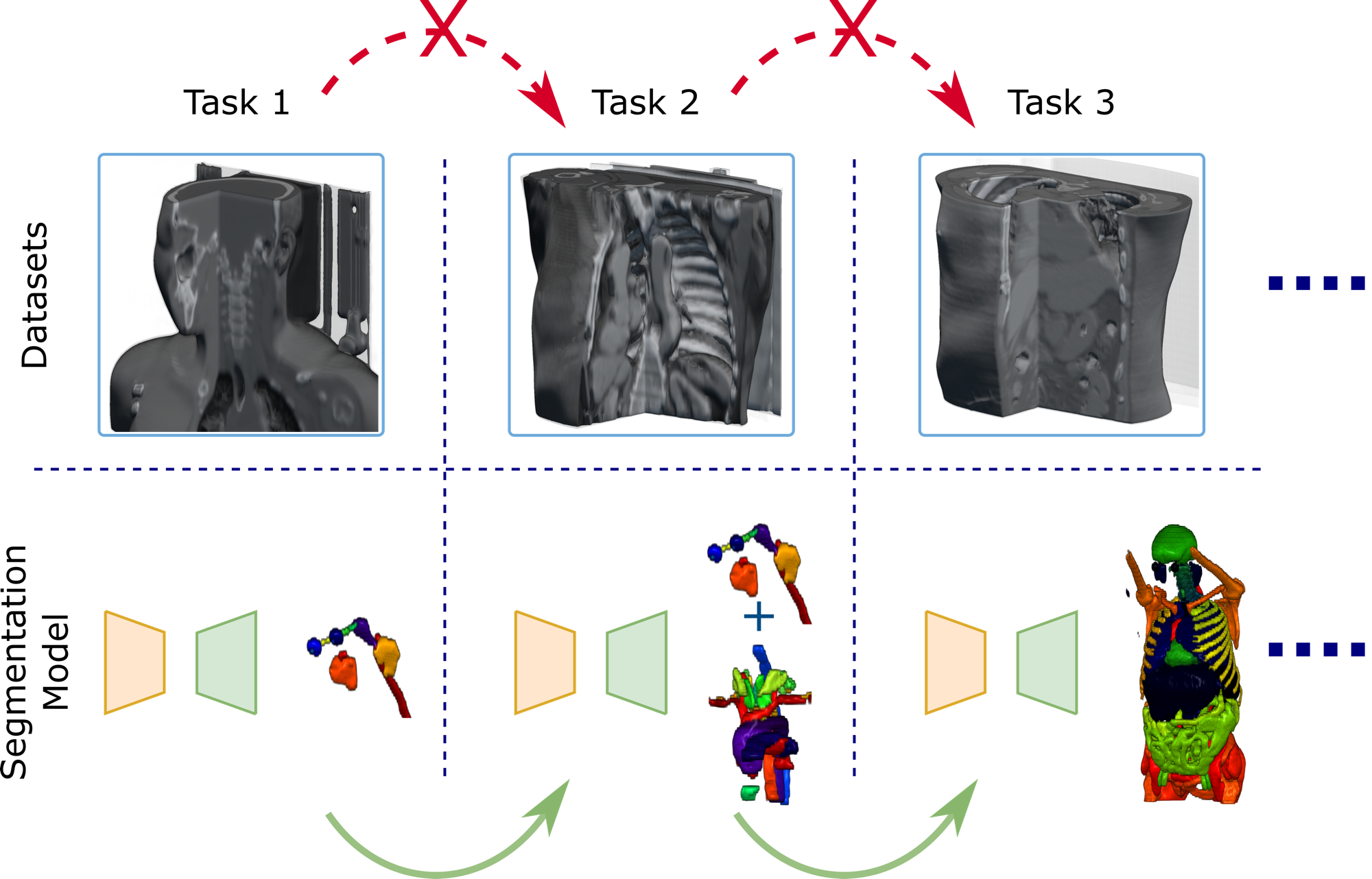}
    \caption{Illustration of the continual multi-organ segmentation. At each continual learning step, only the previously trained model is available (green arrow). Previous datasets are not accessible. We allow organs from different datasets to have overlaps, and these datasets may also contain diseased organs (with tumors).}
    \label{fig:movtivation}
    \vspace{-2mm}
    \end{figure}

\section{Introduction}
Multi-organ segmentation has been extensively studied in medical imaging because of its core importance for many downstream tasks, such as quantitative disease analysis \cite{iyer2016quantitative,ferre2019lymphocyte}, computer-aided diagnosis \cite{roth2015improving,chao2020lymph}, and cancer radiotherapy planning~\cite{jin2021deeptarget,ye2022comprehensive,jin2022towards}. With the emergence of many dedicated labeled organ datasets~\cite{antonelli2022medical} and the fast developments in deep learning segmentation techniques~\cite{isensee2021nnu}, deep segmentation networks trained on specific datasets achieve comparable performance with human observers~\cite{tang2019clinically,ye2022comprehensive,shi2022deep}. However, this setup can have serious limitations in practical deployment for clinical applications. These trained models are pre-trained to segment a fixed number of organs, while in real clinical practice, it is desirable that segmentation models can be dynamically extended to enable segmenting new organs without the (re-)access to previous training datasets or without training from scratch. In this way, patient privacy and data storage issues can be solved, and model development and deployment can be much more efficient. \textcolor{\highlight}{This clinically preferred process can be viewed as continual semantic segmentation (CSS), which is emerging very recently in the natural image domain~\cite{michieli2019incremental,cermelli2020modeling,douillard2021plop,zhang2022representation,michieli2021continual,xiao2023endpoints,cermelli2023comformer} but has been only scarcely studied for medical imaging~\cite{ozdemir2018learn,liu2022learning}. Notably, if all labeled datasets are simultaneously accessible, it simplifies to a federated learning~\cite{rieke2020future,shen2023fedmm} or partial label learning~\cite{fang2020multi,shi2021marginal} problem. However, labeled datasets are always sequentially built over time by annotating different organs of interest according to various clinical tasks.} 

Multi-organ CSS faces several major challenges. First, since old datasets are not accessible when training on the new dataset, deep networks may easily forget the previously learned knowledge if no additional constraints are added, which is the most prominent issue (known as catastrophic forgetting~\cite{thrun1998lifelong,kemker2018measuring}) in continual learning. Second, in contrast to natural image datasets that are often completely labeled~\cite{everingham2015pascal,zhou2017scene}, fully annotated medical image datasets are rare, especially for comprehensive multi-organ datasets. \textcolor{\highlight}{For example, concerning both necessity and cost, labeling 143 organs for all datasets is simply infeasible or impossible.} These partially labeled datasets bring up the label conflict issue (semantic shift of the background class~\cite{cermelli2020modeling}), meaning a labeled organ in dataset-1 may become unlabeled background in dataset-2. Third, domain incremental learning is common in multi-organ CSS, since different datasets may contain overlapped yet ``style-different'' organs. Appropriately tackling these domain gaps is non-trivial. E.g., dataset-1 is made up of healthy subjects with normal esophagus annotated, while dataset-2 is a dedicated esophageal cancer dataset where esophagus with tumor is labeled.

There are several recent CSS work in computer vision~\cite{michieli2019incremental,cermelli2020modeling,douillard2021plop,zhang2022representation,michieli2021continual}. MiB loss is often applied to handle the background-label conflicting issue~\cite{cermelli2020modeling,douillard2021plop}. Regularization-based methods are mostly adopted to reduce the forgetting of old knowledge while learning new classes. However, since network parameters are updated on the training of new classes, it is extremely difficult to achieve high performance on both old and new classes. There are few previous works of CSS in medical imaging~\cite{ozdemir2018learn,liu2022learning}. Ozdemir et al. employed only 9 patients with 2 labels to develop a regularization-based CSS preliminary model~\cite{ozdemir2018learn}. The most recent work~\cite{liu2022learning} used MiB loss and prototype matching to continually segment a small number of 5 abdominal organs focusing only on the abdomen CT. When involving a large number of organs (e.g., $\ge$ 100 classes) affiliated with a variety of body parts, such as in whole-body CT scans for practical considerations, this strategy becomes non-scalable and suffers severe performance degradation (as demonstrated in our experiments later). 


A most recent continual classification work~\cite{wu2022class} has empirically shown that a base classification model trained with a sufficiently large number of classes (e.g., 800) is capable of extracting representative features even for new classes. Hence, freezing most part of its parameters and incrementally fine-tuning the newly added last convolutional block for each new task leads to an almost non-forgetting continual classification model, whose performance is close to the joint learning upper bound for both old and new classes.

Motivated by the observation in continual classification, in this work, we propose a novel architecture-based continual multi-organ segmentation framework. On the basis of the common encoder + decoder architecture of segmentation networks, we demonstrate that its encoder is capable of extracting representative deep features (non-specific to organ or body part) for the new data. Hence, we can freeze the encoder and incrementally add a separate decoder for each new learning task. \textcolor{\highlight}{Under this scheme, when adding a new task, organs learned in previous tasks will never be forgotten because the encoder is frozen, and previous decoders are independent of the new task. In addition, the new decoder is trained separately to segment a fixed number of foreground organs using only the new dataset. Hence, it avoids the background-label conflict with previous datasets during training}. Yet, this scheme can lead to a swelling model as tasks expand. To make it scalable, a progressive trimming method using neural architectural search (NAS) and teacher-student-based knowledge distillation (KD) is exploited to maintain the \textit{overall model complexity} and \textit{inference time} comparable to the original single network. Finally, to merge organ predictions originating from different decoders and incorporate both healthy and pathological organs appearing in different datasets, we propose a body-part and anomaly-aware output merging scheme using automated body part and tumor predictions.

\begin{figure*}[ht!]
    \centering
    \includegraphics[width=0.98\textwidth]{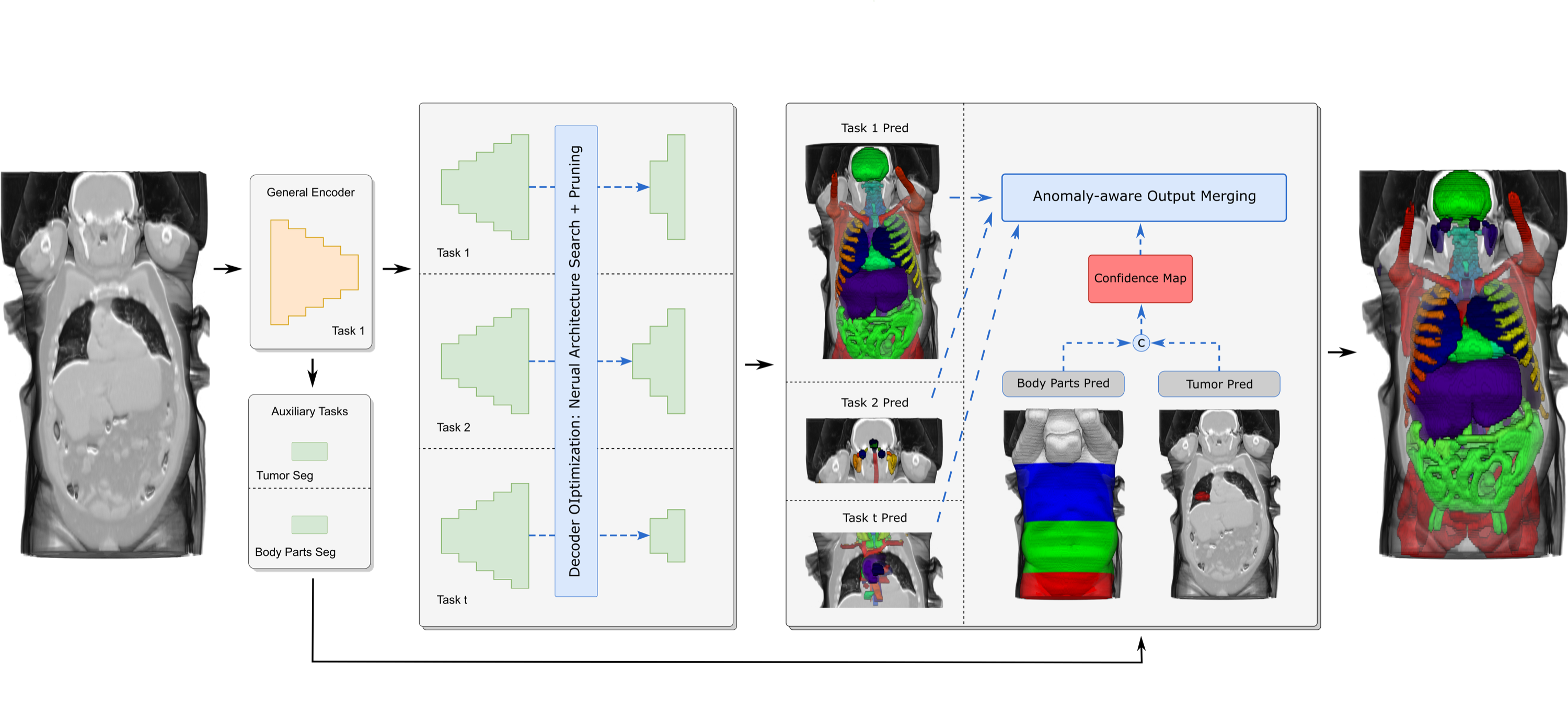}
    \caption{Overall framework of the proposed continual multi-organ segmentation, \textcolor{\edited}{which is composed of a General Encoder, multiple optimized and pruned decoders (one for each learning step), and a body-part and anomaly-ware output merging module. After training the base encoding/decoding segmentation network using $D_1$, the General Encoder is frozen afterward, and separate trainable decoders are incrementally added to continually learn new datasets, which leads to a non-forgetting architecture. Decoder optimization and pruning are applied at each learning step to maintain a reasonable model complexity. Finally, the merging module is designed to combine organs from all decoders.} }\label{fig:workflow}
\end{figure*}

In summary, the main contributions are as follows:
    \begin{itemize}
        \item We are the first to comprehensively study the multi-organ continual semantic segmentation (CSS) problem with a clinically desirable number of organs (143 organs) across different body parts (head \& neck, chest, abdomen) to more sufficiently and efficiently support medical diagnosis and treatment planning purposes.




        \item We propose the first \textcolor{\edited}{(pure)} architecture-based multi-organ continual segmentation framework. Consisting of a general encoder, continually expanded and pruned decoders, and a body-part and anomaly-aware output merging module, the proposed network  avoids the notorious catastrophic forgetting in CSS while being scalable (maintaining the model complexity similar to other types of CSS approaches).
        

        \item Continually trained and validated on 3D CT scans of 2500+ patients compiled from four different datasets, our scalable unified model can segment total of 143 whole-body organs with very high accuracy, closely reaching the upper bound performance level of four well-trained individual models (i.e., nnUNet~\cite{isensee2021nnu}). 
        

    \end{itemize}

\section{Related Work}

\noindent\textbf{Multi-Organ Segmentation.} 
Automated multi-organ segmentation (MOS) is a challenging task in medical imaging with a long study history. The early registration-based atlas approach faces difficulty when large organ variation, tumor growth, or image acquisition differences exist. 
Recently, deep learning-based methods ~\cite{isensee2021nnu,yu2020c2fnas,jin2021deeptarget,guo2020organ,guo2021deepstationing,ji2019scribble,raju2020user} have achieved great success when working on specific datasets with the same set of labeled organs. However, in practice, there are often partially labeled datasets, each with only one or a few labeled organs. Several recent works explore training a joint single model leveraging on multiple partially labeled datasets~\cite{zhou2019prior,fang2020multi,shi2021marginal,petit2021iterative,zhang2021dodnet}. To address the major issue of background label conflicts, the marginal loss is often adopted to merge all unlabeled organs with the background~\cite{fang2020multi,shi2021marginal}. Different from these previous works that require all training datasets to be available/accessible at once, we train a single multi-organ segmentation model incrementally on multi-center partially-labeled datasets, with no access to previous datasets during the sequential process. 

\noindent\textbf{Continual Learning.} 
Continual Learning aims to update a model from a sequence of new tasks and datasets without catastrophic forgetting~\cite{goodfellow2013empirical,kemker2018measuring}. 
There are three main categories~\cite{9349197}. 
\textit{Rehearsal-based} methods store a limited amount of training exemplars from old classes as raw images~\cite{rebuffi2017icarl,hou2019learning,liu2021rmm,chaudhry2019tiny,buzzega2020dark}, embedded features~\cite{hayes2020remind,iscen2020memory} or generators~\cite{ostapenko2019learning,shin2017continual}. However, it may be impracticable in real-world practice when data privacy is concerned, e.g., medical scans across multiple hospital sites are inaccessible. 
\textit{Regularization-based} methods constrain the model plasticity either through regularization on weights~\cite{aljundi2018memory,chaudhry2018riemannian,kirkpatrick2017overcoming,zenke2017continual,kumar2021bayesian} and gradients~\cite{lopez2017gradient,chaudhry2018efficient}, or knowledge distillation on output logits~\cite{li2017learning,schwarz2018progress,rebuffi2017icarl,castro2018end} and intermediate features~\cite{dhar2019learning,douillard2020podnet,zhu2021prototype,zhu2022self}. Although without storing exemplars, they cannot guarantee desirable performance on challenging tasks. 
\textit{Architecture-based} methods aim at either dynamically dividing task-specific partial network~\cite{golkar2019continual,hung2019compacting,mallya2018packnet,serra2018overcoming}, which suffers from running out of trainable parameters or expanding the network by freezing the old model and adding new parameters for new tasks~\cite{rusu2016progressive,li2019learn,wang2017growing,yoon2017lifelong,wu2022class,wang2022learning,ma2022progressive}, which guarantee no-forgetting performance but result in gradually growing/swelling model sizes. 
Our work falls into the expanding category, and we perform network pruning for each new task to control the overall model complexity.


\noindent\textbf{Continual Semantic Segmentation.} 
Continual semantic segmentation (CSS) is an emerging research topic with limited previous studies. 
Besides catastrophic forgetting, CSS faces the same challenge as partially labeled segmentation known as \textit{background shift}~\cite{cermelli2020modeling}. 
ILT~\cite{michieli2019incremental} proposes a CIS setting with a simple knowledge distillation solution. MiB~\cite{cermelli2020modeling} adapts marginal loss for both classification and distillation to solve background shift. A local-pooling-based distillation is applied to intermediate features in PLOP~\cite{douillard2021plop}. CSWKD~\cite{phan2022class} weights the distillation loss based on the old and new class similarity. \textcolor{\edited}{SDR~\cite{michieli2021continual} propose to regularize the latent feature space using prototype matching and contrastive learning}. Other than knowledge distillation, RCIL~\cite{zhang2022representation} designs a two-branch module for decoupling the representation learning of old and new classes. 
In multi-organ segmentation, only one study LISMO~\cite{liu2022learning} applies CSS, based on MiB and prototype matching \textcolor{\edited}{adapted from SDR~\cite{michieli2021continual}}, to segment five abdominal organs, which is an easy setting merely focusing on a single body part (abdomen). Our work is generalized for significantly more organ classes that are located in a large range of body parts (head \& neck, chest, abdomen, hip \& thigh).


\section{Method}


\noindent\textbf{\bf Problem Formulation.} We aim to sequentially and continuously learn a single multi-organ segmentation model from several partially-labeled datasets one by one. Let $D=\{D_1, \dots, D_T\}$ denote a sequence of data. When training on $D_t$, all previous training data $\{D_p, p<t\}$ are not accessible. For the $t^{th}$ dataset $D_t=\{{X_i}^t, {Y_i}^t\}_{i=1}^{n_t}$ with $C_t$ organ classes, let $X^t$ and $Y^t$ denote the input image and the corresponding organ label in the $t^{th}$ dataset, the prediction map for voxel location, $j$, and output class $c^t$:
\begin{align} 
    \hat{Y}^t(j) &= f_d\left(Y^t(j)=c^t|f_e\left(X^t; W_e\right); W_d\right), \\
    \hat{\mathbf{Y}} &=\bigcup_{t=1}^{T}{\hat{Y}^t}, 
    \label{eq:gs}
\end{align} 
where $f_e$, $f_d$, $W_e$, and $W_d$ denote the CNN functions and the corresponding parameters for the encoding and decoding paths, respectively. The final prediction $\hat{\mathbf{Y}}$ is the union (with possible class overlapping) of all previous predictions.

\noindent\textbf{Overall Training Process.} Figure~\ref{fig:workflow} illustrates the proposed multi-organ continual segmentation framework, which is composed of an encoder, multiple optimized and pruned decoders (one for each $D_t$), a body-part, and anomaly-ware output merging module. It starts from training a base encoding/decoding segmentation network using a comprehensive dataset $D_1$. We hypothesize that the well-trained encoder on $D_1$, represented as a General Encoder, is capable of extracting representative features (universal to all organs and datasets) to facilitate the subsequent learning tasks. Hence, this General Encoder is fixed afterward, and separate trainable decoders are incrementally added at the future learning steps, which leads to a non-forgetting architecture. Decoder optimization and pruning are also conducted at each learning step to maintain the model complexity comparable to a single network. Finally, by merging predictions from all decoders, we obtain a single unified model that can segment all organs of interest.

\subsection{General Encoder Training} \label{sec:ge}

Ideally, for whole-body multi-organ segmentation, we expect to construct a sufficiently representative and universal General Encoder that extracts deep image features to capture and encode all visual information inside the full human body. Compared to the image statistics of broad natural image databases, medical images exist in a much more confined semantic domain, i.e., the human body is anatomically structured and composed of distinct body parts, no matter with or without diseases. This makes it feasible to learn a strong universal General Encoder competently capturing the holistic human body CT imaging statistics using large or not-so-limited multi-organ datasets. Sharing a similar idea, a very recent continual classification work~\cite{wu2022class} has empirically shown that a base classification model trained with a sufficiently large number of classes (e.g., 800) in ImageNet is capable of extracting representative features even for new classes. Here, our goal is to build a single unified segmentation model to accurately and continually segment up to 143 whole-body organs in CT scans (appeared in multiple datasets of both healthy subjects and diseased patients). 

To train the General Encoder for multi-organ continual segmentation, we recommend starting with the publicly available TotalSegmentator~\cite{wasserthal2022totalsegmentator} dataset as $D_1$, which consists of 1204 CT scans with a total of 103 labeled whole-body organs. These are routine diagnostic CT scans of different body parts with various scanning protocols. \textcolor{\highlight}{Besides this comprehensive dataset, we also supplement the General Encoder with auxiliary body-part segmentation and abnormal/tumor segmentation tasks. The body part labels can be obtained based on axial CT slice scores predicted by an automated body part regression algorithm~\cite{yan2018unsupervised}. As the slice score is monotonously correlated with the patient's anatomic height, slices with key landmarks can be determined to divide the whole body into four major regions, i.e., head \& neck, chest, abdomen, and hip \& thigh. The abnormal/tumor segmentation head is trained using dedicated tumor datasets. By involving these additional tasks, the General Encoder explicitly recognizes each pixel's anatomy region (body part) and potential abnormal tissues, which may be beneficial for learning better pixel representations. Moreover, the body part and tumor segmentation results can be further utilized in the output merging step to combine outputs from all decoders and reduce potential distal false positives from different decoders. For implementation, light-weighted body parts and tumor segmentation heads are added to the General Encoder using only the FCN8-like projection layers ($0.04\times$ size of a regular decoder)~\cite{jin2019accurate}.} 

\subsection{Decoder Optimization \& Pruning}

\textcolor{\highlight}{As the continual segmentation step extends, the proposed model complexity may escalate. Therefore, after initially training the decoder at each continual step, we further apply a progressive optimization and pruning procedure to scale down the decoder complexity with the least possible performance drop.}


\noindent\textbf{\bf Decoder Optimization via Neural Architectural Search (NAS).} 
We first conduct NAS to optimize the decoder's segmentation performance and possibly reduce the decoder's parameters.  Let $\phi\left( \cdot; \omega_{x \times y \times z} \right)$ denote a composition function of consecutive operations: batch normalization, a rectified linear unit, and a convolution layer with an $x \times y \times z$ dimension kernel. Inspired by previous work~\cite{zhu2019v, guo2020organ}, different convolutional layers may require various 2D/3D kernel types to segment 3D organs. Hence, we search for a set of possible convolutional kernels tailored to our problem: projection convolution $\phi\left( \cdot; \omega_{1 \times 1 \times 1} \right)$, 2D convolution $\phi\left( \cdot; \omega_{3 \times 3 \times 1} \right)$, pseudo-3D (P3D) convolution $\phi\left( \phi\left( \cdot; \omega_{3 \times 3 \times 1} \right); \omega_{1 \times 1 \times 3} \right)$, and 3D convolution $\phi\left( \cdot; \omega_{3 \times 3 \times 3} \right)$. To simplify the searching process, we use only one type of convolutional kernel to build each decoding block.  At the end of the search, we determine the architecture of each block by choosing the $\phi$ corresponding to the largest weight value. Besides the optimized decoder performance, the searched 2D and P3D kernel parameters are only 1/3 and 4/9 of the 3D one, which also trims down the network parameter numbers.
\begin{figure}
    \centering
    \includegraphics[width=0.95\columnwidth]{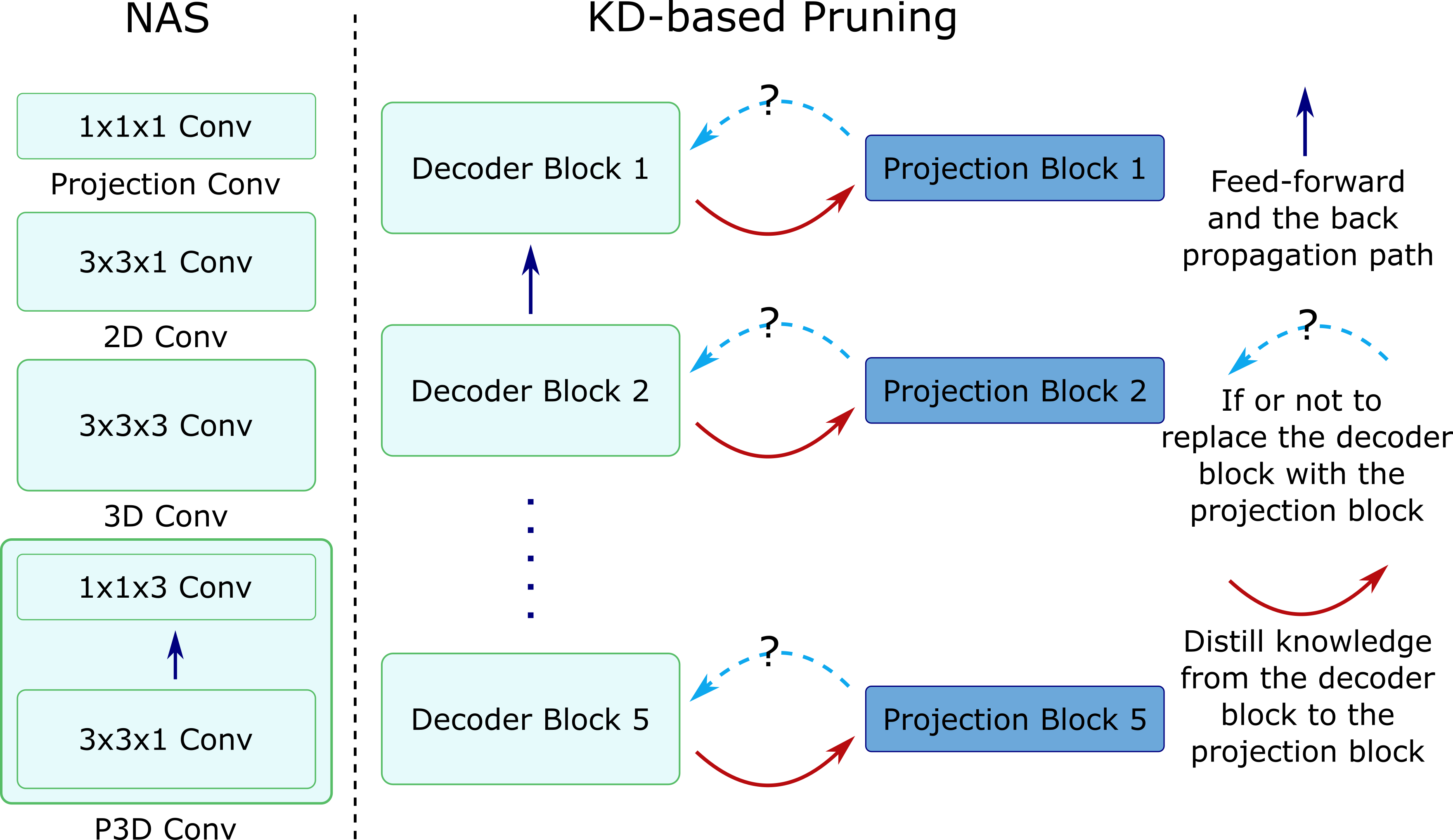}\label{fig:trim}
    \caption{Illustration of the decoder optimization and pruning via neural architectural search and  knowledge distillation. }\label{fig:pruning}
\end{figure}

\noindent\textbf{\bf Decoder Compression via Knowledge Distillation (KD).} After NAS, we further prune the decoder by designing a convolution block-wise teacher-student-based Knowledge Distillation (KD) method. 
Each convolutional block is fixed and used as the teacher block. Next, we pair each teacher block with a projection block (i.e., a convolutional block with projection layers with kernel size 1), aiming to replace the teacher block with this projection block. The mean-square error loss is adopted to match the feature maps of the teacher block to the student block. Note that the student blocks have no path connection (hence no gradient back-propagation). To reduce the optimization difficulty, the deeper level of the decoding blocks is optimized first. Once the KD training of the deeper blocks is saturated, we freeze them and progressively move to the shallower ones. Figure~\ref{fig:pruning} illustrates the pruning method. After this process, there are $2^5$ decoding paths when choosing between the original and the projection convolutional block, where all possible combinations are enumerated, and the corresponding segmentation performance and decoding parameter numbers are recorded.  We use the decreased segmentation Dice score (\%) to select the most possibly pruned decoding path. This decreased Dice score is defined by a performance drop tolerance parameter $\tau$. In ablation experiments, we use $\tau \in \{1\%, 3\%, 5\%\}$ to inspect the model compression results. The final results are reported using $\tau=1\%$. For the detailed distillation training process, please refer to the supplementary materials. 

\subsection{Body-part \& Anomaly-aware Output Merging} 
\label{sec:merging}

\textcolor{\highlight}{We exploit the body part and anomaly predictions from two auxiliary tasks and propose a simple yet effective rule-based approach to combine the predictions from all decoders. Specifically, for each dataset/task, we pre-compute the merged bounding boxes of all labeled organs. Next, we calculate the average body part distribution map $P^t$ for each dataset $t$ by overlapping the averaged bounding box to the body part labels. Let $\hat{Y}^{\epsilon}$ denote the distinct tumor prediction, $\odot$ denote the element-wise multiplication, and $J$ denote the matrix of ones}, the weighting map $M^t$ is calculated using Eq.~(\ref{eq:weighting}), i.e., only when $\hat{Y}^{\epsilon} \rightarrow 0$ and $P^{t} \rightarrow 1$ s.t. the $M^t \rightarrow 1$, whereas $M^t \rightarrow 0.5$ for the rest states. We use the entropy function Eq.~(\ref{eq:confidence}) to compute the confidence map.

\begin{align}
M^{t} &= J - \frac{1}{2} \left(J - P^t + \hat{Y}^{\epsilon} \odot P^t\right) \label{eq:weighting}\\
H^t &= - \left(M^{t} \odot \hat{Y}^t\right) \log\left(M^{t} \odot \hat{Y}^t\right), \label{eq:confidence}\\
\mathbf{H}(j) &= \bigcup_{\forall \hat{Y}(j)^t \neq 0}{H^t(j)}, t\in\{1, \dots, T\},\\
\hat{\mathbf{Y}}(j) &= \hat{Y}^{\argmin(\mathbf{H}(j))}(j)\label{eq:final}
\end{align}

For each voxel, we collect a set $\mathbf{H}(j)$, for all $\hat{Y}(j)^t \neq 0$. Depicted in Eq.~(\ref{eq:final}), the final output class $\hat{\mathbf{Y}}(j)$ is determined using the prediction $\hat{Y}^t(j)$, of which with the smallest $H^t(j)$. For the detailed merging setups, 
\textcolor{\edited}{please refer to the supplementary Sec.~\ref{S-supp:implement}}.

\section{Experiments}





\begin{table*}[t]
\centering
\caption{Continual multi-organ segmentation final results on two orders of our datasets. Dataset names are followed by their class numbers. Mean DSC (\%, $\uparrow$), HD95 (mm, $\downarrow$) and ASD (mm, $\downarrow$) are evaluated on each dataset as well as all classes (All). `Params \#': decoder(s) parameter number of the final model (\# (MB)) and the relative number (Rel \#) compared to the original nnUNet decoder. 
$\dagger$: ILT is reimplemented using a frozen encoder setting and the unbiased loss from MiB for better performance.}
\label{tab:main}
\vspace{2mm}
\resizebox{\textwidth}{!}{%
\begin{tabular}{lrrrrrrrrrrrrrrrrr}
\hline
                                          & \multicolumn{3}{c}{\textbf{TotalSeg (103)}}                                                               & \multicolumn{3}{c}{\textbf{ChestOrgan (31)}}                                                              & \multicolumn{3}{c}{\textbf{HNOrgan (13)}}                                                                 & \multicolumn{3}{c}{\textbf{EsoOrgan (1)}}                                                                & \multicolumn{3}{c}{\textbf{All (143)}}                                                                    & \multicolumn{2}{c}{\textbf{Params \#}}                                          \\ \cline{2-18} 
\multirow{-2}{*}{\textbf{Methods}}        & \multicolumn{1}{l}{\textbf{DSC}} & \multicolumn{1}{l}{\textbf{HD95}} & \multicolumn{1}{l|}{\textbf{ASD}}  & \multicolumn{1}{l}{\textbf{DSC}} & \multicolumn{1}{l}{\textbf{HD95}} & \multicolumn{1}{l|}{\textbf{ASD}}  & \multicolumn{1}{l}{\textbf{DSC}} & \multicolumn{1}{l}{\textbf{HD95}} & \multicolumn{1}{l|}{\textbf{ASD}}  & \multicolumn{1}{l}{\textbf{DSC}} & \multicolumn{1}{l}{\textbf{HD95}} & \multicolumn{1}{l|}{\textbf{ASD}}  & \multicolumn{1}{l}{\textbf{DSC}} & \multicolumn{1}{l}{\textbf{HD95}} & \multicolumn{1}{l|}{\textbf{ASD}}  & \multicolumn{1}{l}{\textbf{\# (MB)}} & \multicolumn{1}{l}{\textbf{Rel \#}} \\ \hline
                                          & \multicolumn{15}{c}{\textbf{Order A: TotalSeg $\rightarrow$ ChestOrgan $\rightarrow$ HNOrgan $\rightarrow$ EsoOrgan}}                                                                                                                                                                                                                                                                                                                                                                                                                   & \multicolumn{2}{l}{}                                                            \\ \hline
\multicolumn{1}{l|}{\textbf{MiB}~\cite{cermelli2020modeling}}         & 7.65                             & 119.66                            & \multicolumn{1}{r|}{67.41}         & 19.24                            & 37.14                             & \multicolumn{1}{r|}{8.34}          & 6.37                             & 7.40                              & \multicolumn{1}{r|}{2.38}          & 86.92                            & 4.33                              & \multicolumn{1}{r|}{1.09}          & 8.51                             & 98.98                             & \multicolumn{1}{r|}{51.98}         &                                      &                                          \\
\multicolumn{1}{l|}{$\textbf{ILT}^\dagger$~\cite{michieli2019incremental}} & 10.87                            & 192.23                            & \multicolumn{1}{r|}{116.20}        & 27.87                            & 36.93                             & \multicolumn{1}{r|}{7.41}          & 6.39                             & 4.04                              & \multicolumn{1}{r|}{0.81}          & 85.75                            & 4.57                              & \multicolumn{1}{r|}{1.17}          & 11.99                            & 148.96                            & \multicolumn{1}{r|}{86.34}         &                                      &                                          \\
\multicolumn{1}{l|}{\textbf{PLOP}~\cite{douillard2021plop}}        & 37.30                            & 53.71                             & \multicolumn{1}{r|}{23.33}         & 51.74                            & 35.36                             & \multicolumn{1}{r|}{8.71}          & 25.38                            & 16.12                             & \multicolumn{1}{r|}{9.24}          & 82.90                            & 6.21                              & \multicolumn{1}{r|}{1.62}          & 39.01                            & 46.63                             & \multicolumn{1}{r|}{18.48}         &                                      &                                          \\
\multicolumn{1}{l|}{\textbf{LISMO}~\cite{liu2022learning}}       & 10.82                            & 129.82                            & \multicolumn{1}{r|}{76.92}         & 28.24                            & 36.33                             & \multicolumn{1}{r|}{9.08}          & 6.30                             & 12.93                             & \multicolumn{1}{r|}{4.14}          & \textbf{87.12}                   & \textbf{4.24}                              & \multicolumn{1}{r|}{\textbf{1.05}}          & 12.11                            & 96.89                             & \multicolumn{1}{r|}{54.71}         &                                      &                                          \\
\multicolumn{1}{l|}{\textbf{RCIL}~\cite{zhang2022representation}}       & 42.58                            & 48.28                            & \multicolumn{1}{r|}{23.24}         & 57.76                            & 33.95                             & \multicolumn{1}{r|}{9.12}          & 27.96                             & 16.88                             & \multicolumn{1}{r|}{8.59}          & 84.72                   & 5.95                              & \multicolumn{1}{r|}{1.16}          & 42.43                            & 44.89                             & \multicolumn{1}{r|}{18.67}         & \multirow{-5}{*}{15.068}             & \multirow{-5}{*}{1.00}                   \\ \hline
\multicolumn{1}{l|}{\textbf{Ours}}        & \textbf{92.98}                   & \textbf{4.09}                     & \multicolumn{1}{r|}{\textbf{0.98}} & \textbf{78.26}                   & \textbf{9.17}                     & \multicolumn{1}{r|}{\textbf{1.82}}          & \textbf{83.97}                   & \textbf{2.22}                     & \multicolumn{1}{r|}{\textbf{0.59}} & 86.94                            & 5.04                              & \multicolumn{1}{r|}{1.11}          & \textbf{88.74}                   & \textbf{5.28}                     & \multicolumn{1}{r|}{\textbf{1.14}} & 14.669                               & 0.98                                     \\ \hline
                                          & \multicolumn{15}{c}{\textbf{Order B: TotalSeg $\rightarrow$ HNOrgan $\rightarrow$ ChestOrgan $\rightarrow$ EsoOrgan}}                                                                                                                                                                                                                                                                                                                                                                                                                   & \multicolumn{1}{l}{}                 & \multicolumn{1}{l}{}                     \\ \hline
\multicolumn{1}{l|}{\textbf{MiB}~\cite{cermelli2020modeling}}         & 10.35                            & 136.77                            & \multicolumn{1}{r|}{63.51}         & 65.63                            & 14.37                             & \multicolumn{1}{r|}{1.94}          & 6.29                             & 24.83                             & \multicolumn{1}{r|}{7.22}          & 86.79                            & 4.31                              & \multicolumn{1}{r|}{1.08}          & 20.00                            & 68.82                             & \multicolumn{1}{r|}{29.87}         &                                      &                                          \\
\multicolumn{1}{l|}{$\textbf{ILT}^\dagger$~\cite{michieli2019incremental}} & 13.12                            & 201.66                            & \multicolumn{1}{r|}{106.51}        & 67.28                            & 14.21                             & \multicolumn{1}{r|}{1.88}          & 6.18                             & 3.12                              & \multicolumn{1}{r|}{0.95}          & 85.52                            & 4.80                              & \multicolumn{1}{r|}{1.25}          & 22.31                            & 115.23                            & \multicolumn{1}{r|}{59.34}         &                                      &                                          \\
\multicolumn{1}{l|}{\textbf{PLOP}~\cite{douillard2021plop}}        & 30.82                            & 62.07                             & \multicolumn{1}{r|}{23.14}         & 70.18                            & 13.05                             & \multicolumn{1}{r|}{2.36}          & 15.77                            & 11.09                             & \multicolumn{1}{r|}{3.84}          & 83.41                            & 6.11                              & \multicolumn{1}{r|}{1.54}          & 36.49                            & 44.78                             & \multicolumn{1}{r|}{16.01}         &                                      &                                          \\
\multicolumn{1}{l|}{\textbf{LISMO}~\cite{liu2022learning}}       & 14.04                            & 90.17                             & \multicolumn{1}{r|}{47.81}         & 67.19                            & 14.88                             & \multicolumn{1}{r|}{1.93} & 6.15                             & 9.13                              & \multicolumn{1}{r|}{1.44}          & 86.87                            & \textbf{4.18}                     & \multicolumn{1}{r|}{\textbf{1.03}} & 22.92                            & 57.71                             & \multicolumn{1}{r|}{28.22}         &             &                    \\ 
\multicolumn{1}{l|}{\textbf{RCIL}~\cite{zhang2022representation}}       & 35.24                            & 59.81                            & \multicolumn{1}{r|}{24.20}         & 70.74                            & 12.98                             & \multicolumn{1}{r|}{2.22}          & 18.43                             & 11.81                             & \multicolumn{1}{r|}{3.65}          & 84.17                   & 6.14                              & \multicolumn{1}{r|}{1.09}          & 39.85                            & 45.52                             & \multicolumn{1}{r|}{15.07}         & \multirow{-5}{*}{15.068}             & \multirow{-5}{*}{1.00}                   \\ \hline
\multicolumn{1}{l|}{\textbf{Ours}}        & \textbf{92.98}                   & \textbf{4.09}                     & \multicolumn{1}{r|}{\textbf{0.98}} & \textbf{78.26}                   & \textbf{9.17}                     & \multicolumn{1}{r|}{\textbf{1.82}}          & \textbf{83.97}                   & \textbf{2.22}                     & \multicolumn{1}{r|}{\textbf{0.59}} & \textbf{86.94}                            & 5.04                              & \multicolumn{1}{r|}{1.11}          & \textbf{88.74}                   & \textbf{5.28}                     & \multicolumn{1}{r|}{\textbf{1.14}} & 14.669                               & 0.98                                     \\ \hline\hline
\multicolumn{1}{l|}{\textbf{Upper bound}} & 93.24                            & 3.29                              & \multicolumn{1}{r|}{0.83}          & 78.45                            & 8.16                              & \multicolumn{1}{r|}{1.83}          & 84.35                            & 2.38                              & \multicolumn{1}{r|}{0.60}          & 87.15                            & 4.44                              & \multicolumn{1}{r|}{0.98}          & 89.02                               & 4.41                                & \multicolumn{1}{r|}{1.06}            & 15.07$\times$4                               & 1.0$\times$4                                     \\ \hline
\end{tabular}%
}
\end{table*}


\begin{figure*}
    \centering
    \includegraphics[trim={0 0.6em 0 0.6em}, clip, width=0.40\paperwidth]{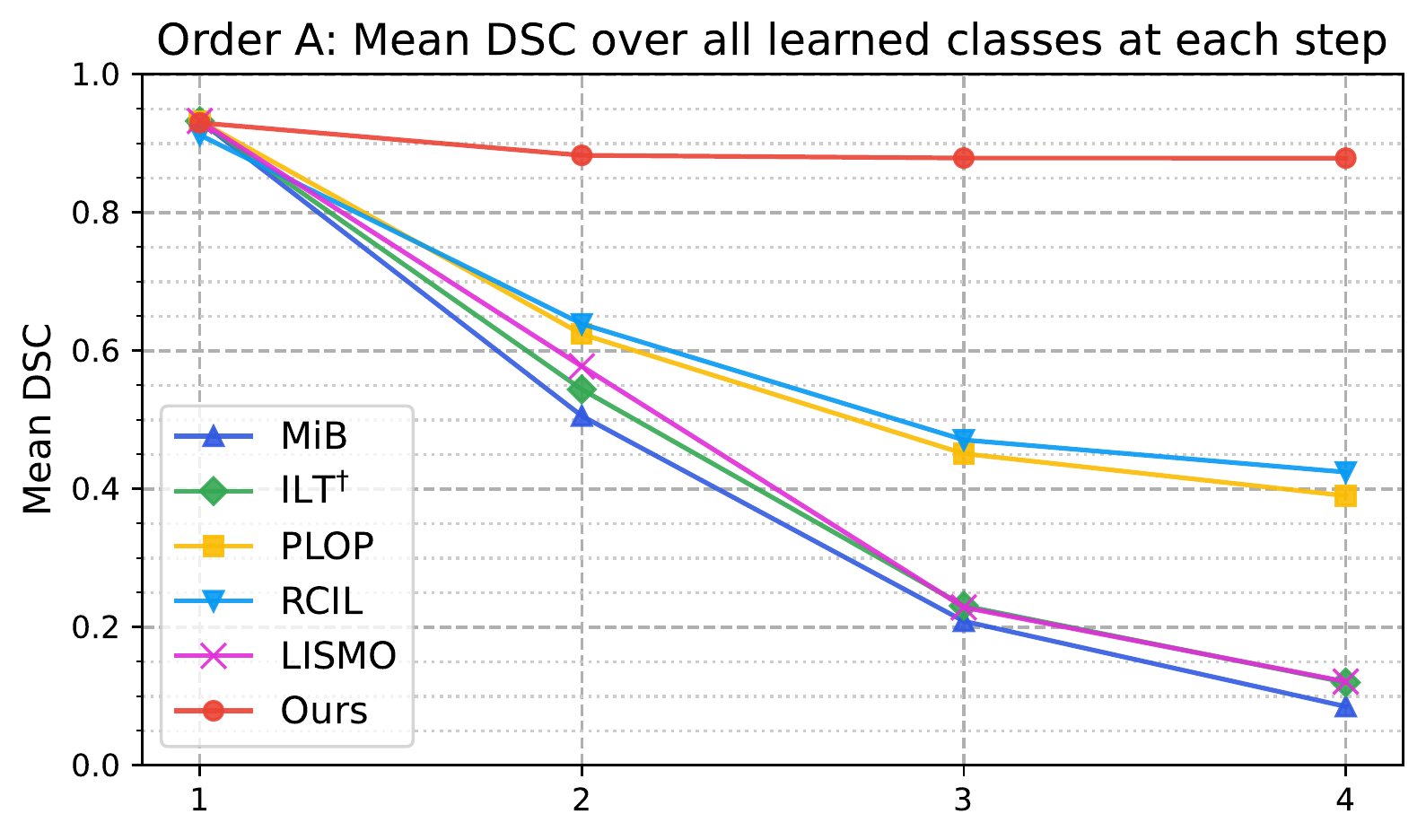}
    \includegraphics[trim={0 0.6em 0 0.6em}, clip, width=0.40\paperwidth]{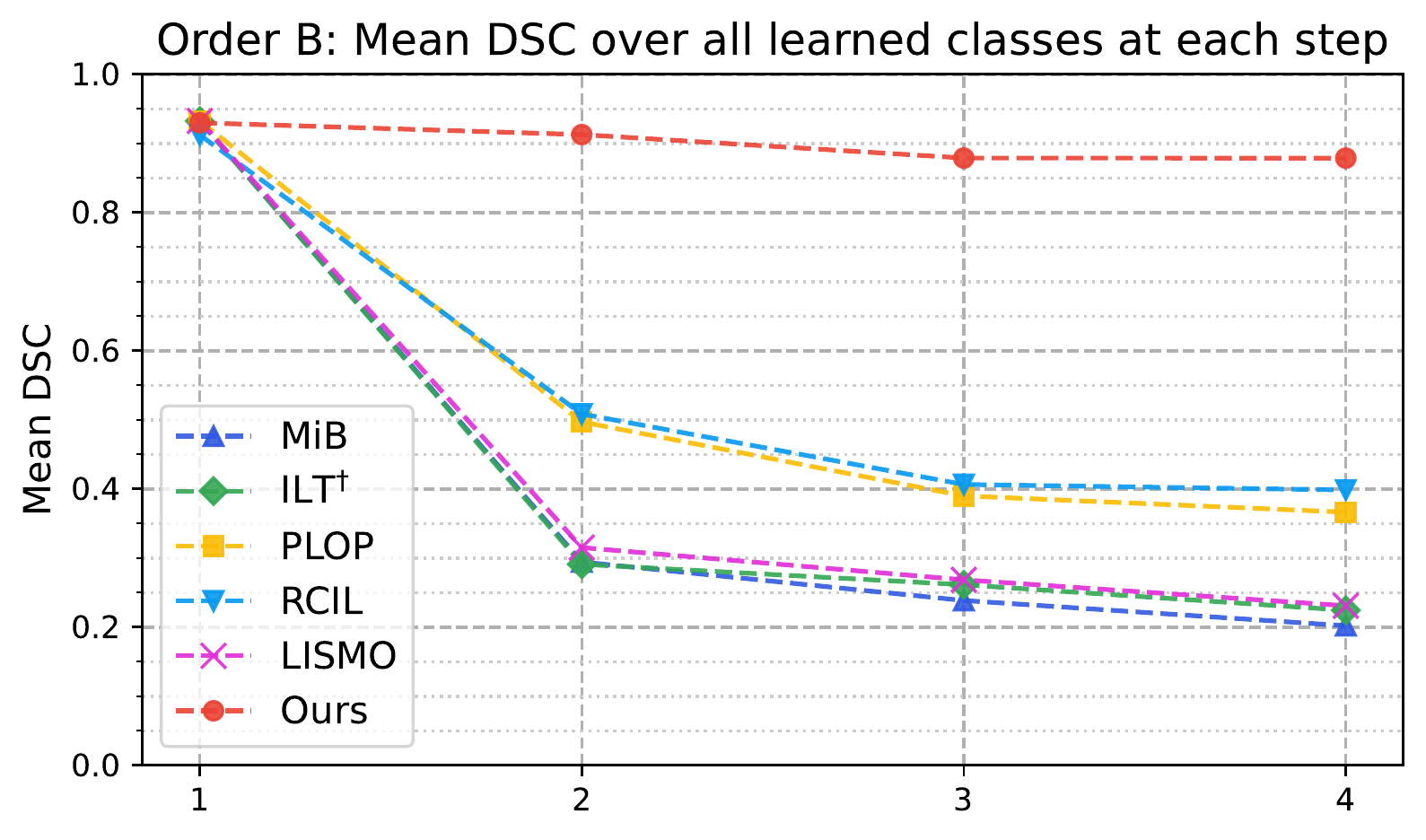}
    \vspace{1mm}
    \caption{The mean DSCs over all learned classes at each step of {\bf Order A} (left, solid line) and {\bf Order B} (right, dashed line).}\label{fig:step_all}
\end{figure*}


{\noindent \bf Datasets:} We evaluated our method using 2500+ patients from one public and three private partially labeled multi-organ datasets. TotalSegmentator~\cite{wasserthal2022totalsegmentator} consists of 1204 CT scans of different body parts with a total of 103 labeled anatomical structures (26 major organs, 59 bone instances, 10 muscles, and 8 vessels). Note that the face label is removed as it is an artificially created label for patient de-identification purposes after blurring the facial area. In the in-house collection, the ChestOrgan dataset contains 292 chest CT scans, most of which come from early esophageal or lung cancer patients. For the ChestOrgan dataset, 31 chest anatomical structures are labeled, among which 4 overlapped with organs in TotalSegmentator (esophagus, trachea, SVC, pulmonary artery). Another dataset includes 447 head \& neck CT scans (denoted as HNOrgan dataset), where 13 organs are annotated as organs at risk (OARs) in radiation therapy and do not have class overlap with all other datasets. The fourth dataset is a dedicated cancer dataset validating the domain change of CSS, containing 640 diagnostic CT scans of advanced esophageal cancer patients where only the esophagus is labeled (denoted as the EsoOrgan dataset). For the detailed organ list, please refer to the supplementary Sec.~\ref{S-supp:dataset}. By combining all datasets, we have a total of $103+27+13=143$ organ classes from 2583 unique patients. For each of these four datasets, $20\%$ is randomly chosen as an independent testing set, while the rest is used as training + validation in each continual learning step.  

In addition, for the purpose of training and validating our abnormality segmentation module, we further collect CT scans from 304 esophageal (private) and 625 lung cancer (public with labels) patients where the 3D tumor masks are segmented. 


\noindent{\bf Overall CSS Training Process:} \textcolor{\highlight}{In our CSS experiment, the model is trained to segment organs sequentially at multiple steps. At each step $t$, the model is trained on the specific dataset $D_t$ without access to any other datasets. Specifically, at step-1, $D_1$ is first used to train both the General Encoder and the associated decoder, where the decoder is further optimized and pruned using $D_1$. After that, $D_1$ cannot be accessed in any future steps. This process is repeated for step 2, $...$, $T$, except that at each step-$t$, $D_t$ is only used to train, optimize and prune $D_t$ dedicated decoder keeping the General Encoder always frozen.}

{\noindent \bf CSS Protocols:}  We examine two CSS orders with four learning steps. \textcolor{\highlight}{\textbf{Order A} goes as: \textit{TotalSegmentator $\rightarrow$ ChestOrgan $\rightarrow$ HNOrgan $\rightarrow$ EsoOrgan}. \textbf{Order B} goes as: \textit{TotalSegmentator $\rightarrow$ HNOrgan $\rightarrow$ ChestOrgan $\rightarrow$ EsoOrgan}, which exchanges the \textit{ChestOrgan} with \textit{HNOrgan} to demonstrate the effect of different body parts in CSS. All methods (including ours) are trained and evaluated in both orders. To report the final results in CSS setting, we compute segmentation metrics after the last learning step for all the previous datasets.  For reporting the results in any intermediate step $t$, these metrics are calculated after the learning step $t$ for all the datasets $i \le t$. }


{\noindent \bf Metrics:} We report the Dice similarity coefficient (DSC), $95\%$ Hausdorff distance (HD95) and average surface distance (ASD) to quantify the organ segmentation results.



\subsection{Implementation Details}
A $[-1024, 1024]$ HU CT windowing is applied to every CT image. We resample all CT scans to the same resolution: $0.75\times 0.75\times 3.0$mm. The ratio between the training and validation set is 4:1.  ``3d-full res'' version (+ ``moreDA'' data augmentation) of nnUNet~\cite{isensee2021nnu} with DSC+CE losses is adopted for all model training with a batch size of 2. The training patch size is $128\times 128 \times 64$.  \textcolor{\highlight}{We set 8000 epochs for training General Encoder and the associated decoder using the TotalSegmentator dataset in step-1, and 1000 epochs for training the dedicated decoder at each future step-$t$.}


\noindent\textbf{NAS Setting:} \textcolor{\highlight}{At a learning step $t, 1 \le t \le T$, after initially training the decoder, we further exploit NAS to search for the optimal network architecture for the associate decoder. For NAS training, the dataset $D_t$ (training+validation) is initially divided into 1) $60\%$ for network training, 2) $30\%$ for NAS training, and 3) $10\%$ for validation evaluation. The initial learning rate is set to 0.01. We first fix the NAS parameters and train the network for 400 epochs. Then we alternatively update the NAS and network parameters for additional 600 epochs. The batch size is set to 4 in NAS training. Only the validation set is used for updating NAS parameters. After NAS training, we follow the same `moreDA' data augmentation scheme and retrain the searched decoding architecture from scratch using $D_t$ (training+validation) with a re-divided `training-validation' ratio of 4:1.}

\noindent\textbf{Pruning Setting:} \textcolor{\highlight}{After NAS, we perform a block-wise teacher-student KD to compress the decoder by replacing the searched convolutional kernels with the projection kernels. The initial learning rate is 0.01. We fix the teacher networks and train the paired student network for another 500 epochs. MSE loss is used for teacher-student feature map matching. After the pruning is completed, we replace the selected teacher blocks with the student blocks and finetune the trimmed network for 500 epochs with a learning rate of 0.001.}
All models are developed using PyTorch and trained on one NVIDIA A100 GPU. \textcolor{\edited}{Please refer to the supplementary Sec.~\ref{S-supp:implement} for more implementation details.}



%
{\noindent \bf Comparing Methods:} \textcolor{\edited}{We compare our method with five latest leading CSS works, including four regularization-based methods (ILT~\cite{michieli2019incremental}, MiB~\cite{cermelli2020modeling}, PLOP~\cite{douillard2021plop}, RCIL~\cite{zhang2022representation}) and a hybrid of regularization and rehearsal-based method (LISMO~\cite{liu2022learning})}. To ensure comparisons' fairness, we re-implement ILT, MiB, LISMO, and PLOP in the nnUNet framework to guarantee consistent data pre-processing and data augmentation (Re-implementation details are fully disclosed in the supplementary Sec.~\ref{S-supp:impl_comp}). Noted that all four datasets in our experiment are partially labeled, hence, it is not straightforward to compute the upper bound performance using a single model. In this work,  we train a separate nnUNet~\cite{isensee2021nnu} model for each dataset, the results of which can serve as the CSS performance upper bound for each dataset.



\begin{table}
\centering
\caption{Segmentation performance under two 1-step continual learning scenarios with and without freezing the General Encoder. Mean DSC (\%, $\uparrow$), HD95 (mm, $\downarrow$) and ASD (mm, $\downarrow$) are evaluated.}
\vspace{3mm}
\label{tab:abl_unfrz}
\resizebox{\columnwidth}{!}{%
\begin{tabular}{l|crrr|crrr}
\hline
\multirow{3}{*}{\textbf{Settings}} & \multicolumn{4}{c|}{\textbf{TotalSeg $\rightarrow$ ChestOrgan}}                                                                               & \multicolumn{4}{c}{\textbf{TotalSeg $\rightarrow$ HNOrgan}}                                                                                  \\
                                   & \multicolumn{2}{c|}{\textbf{TotalSeg}}                                & \multicolumn{2}{c|}{\textbf{ChestOrgan}}                              & \multicolumn{2}{c|}{\textbf{TotalSeg}}                                & \multicolumn{2}{c}{\textbf{HNOrgan}}                                 \\
                                   & \multicolumn{1}{l}{\textbf{DSC}} & \multicolumn{1}{l|}{\textbf{HD95}} & \multicolumn{1}{l}{\textbf{DSC}} & \multicolumn{1}{l|}{\textbf{HD95}} & \multicolumn{1}{l}{\textbf{DSC}} & \multicolumn{1}{l|}{\textbf{HD95}} & \multicolumn{1}{l}{\textbf{DSC}} & \multicolumn{1}{l}{\textbf{HD95}} \\ \hline
\textbf{unfreezing}                & \multicolumn{1}{r}{51.42}        & \multicolumn{1}{r|}{26.52}         & 78.45                            & 8.16                               & \multicolumn{1}{r}{2.90}         & \multicolumn{1}{r|}{162.09}        & 84.35                            & 2.38                              \\
\textbf{freezing}                  & \multicolumn{1}{r}{92.98}        & \multicolumn{1}{r|}{4.09}          & 77.91                            & 8.37                               & \multicolumn{1}{r}{92.98}        & \multicolumn{1}{r|}{4.09}          & 84.14                            & 2.39                              \\ \hline
\end{tabular}%
}
\vspace{-1em}
\end{table} 

\subsection{Comparison to Leading CSS Methods}

\begin{figure*}[ht!]
    \centering
    \includegraphics[trim={3em 0 0 0}, clip, width=1.0\textwidth]{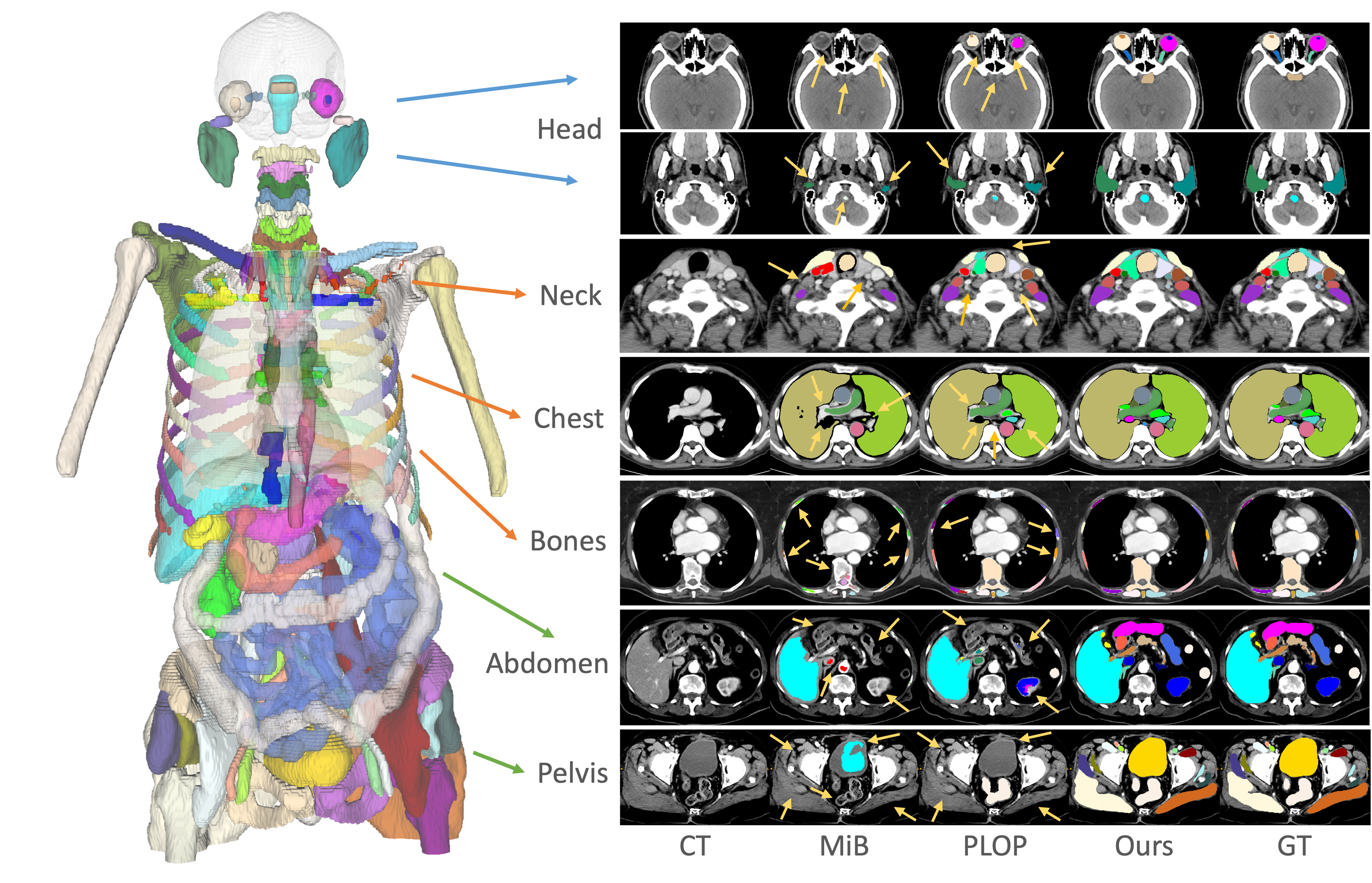}
    \caption{The qualitative comparison of our method with MiB~\cite{cermelli2020modeling} and PLOP~\cite{douillard2021plop}. All the segmentation results are from the last step of CSS order A. Seven quality examples are shown covering different body parts and organ groups. Yellow arrows indicate missing/wrong predictions. MiB and PLOP experience severe forgetting in the head, abdomen and pelvis regions, since these body parts only appear in one or two tasks/datasets; while less forgetting is observed in neck and upper chest regions, which appear in all four tasks/datasets (more suitable for CSS by the MiB and PLOP methods). (For visualization purpose, not all the organs are shown in each example.)}\label{fig:quality}
    \vspace{-1em}
\end{figure*}

\noindent\textbf{Overall Performance:} Table~\ref{tab:main} and  Figure~\ref{fig:step_all} show final segmentation results after continually learning on two orders (each with four steps) of our datasets. Our proposed method significantly outperforms other leading methods on the previously learned three datasets as well as the total 143 organs in both CSS orders. \textcolor{\edited}{The second best performing method RCIL~\cite{zhang2022representation} still experiences catastrophic forgetting and has a mean DSC of 41.43\% and 39.85\% in CSS order A and B, far less than our mean DSC of 88.74\%. Similar performance gaps are noticed on HD95 and ASD metrics (e.g., $\sim$45mm vs. $\sim$5mm in terms of HD95).} Our proposed method achieves very similar performance to the upper bound with a 0.28\% marginal decrease in DSC and a 0.08mm increase in ASD (see Figure~\ref{fig:quality} for qualitative results). In model complexity, the overall parameter number of our four pruned decoders (14.7 MB) is 98\% of an original nnUNet decoder (15.1 MB), which is only 24\% size of the decoders required by achieving the upper bound performance. The running time of the proposed framework (segmenting 143 organs) is slightly longer (+$12\%$) than the running time of a single nnUNet to segment 103 organs. \textcolor{\edited}{For our detailed results of individual organs or organ groups, please refer to the supplementary Sec.~\ref{S-supp:organ}.}

\noindent\textbf{Two CSS Orders:} \textcolor{\highlight}{Table~\ref{tab:main} also demonstrates the segmentation results under two CSS orders (order A and B). Because the proposed framework consists of a frozen General Encoder, independent decoders (each for one continual learning step), and a unified output merging module, our method is order invariant if the base dataset for training General Encoder is the same. On the other hand, the continual learning order may significantly affects the comparison methods. E.g., LISMO has a mean DSC of 28.24\% v.s. 67.19\% on ChestOrgan dataset in order A and B, respectively. }

\textcolor{\highlight}{The significant performance drop of the comparing methods could be caused 
\textcolor{\edited}{by the catastrophic forgetting induced from the bodypart-related domain gap.}
\textcolor{\edited}{In our experiments, we observe that the comparing methods generally work well if new and old datasets share similar domains/body-parts, e.g. ChestOrgan $\rightarrow$ EsoOrgan (second from the left plot of bottom row in the supplementary Figure~C.1)}. However, in whole-body organ continual segmentation, different datasets \textcolor{\edited}{may} cover various body parts with limited overlaps, which causes a large gap in the image domain and significantly deteriorates the performance\textcolor{\edited}{, e.g. ChestOrgan $\rightarrow$ HNOrgan (third from the left plot of bottom row in the supplementary Figure~C.1)}. In contrast, when learning new tasks, our framework keeps previously learned parameters unchanged and avoids knowledge forgetting.} \textcolor{\edited}{Please refer to the supplementary Sec.~\ref{S-supp:ablation} for more detailed results and discussion on the step-wise results of our method and comparing methods.} 

\begin{table}
\centering
\caption{Multi-organ segmentation results using decoder optimization \& pruning. We report the number of decoder parameters and the relative size percentage compared to the original nnUNet decoder when the DSC (\%) is dropped by $\tau \in \{1\%, 3\%, 5\%\}$.
\label{tab:trim}}
\vspace{2mm}
\scalebox{0.8}{
\begin{tabular}{lllll}
\multicolumn{2}{l}{\multirow{2}{*}{\textbf{}}}                                         & \multicolumn{3}{c}{\textbf{DSC Drop}}     \\ \cline{3-5} 
\multicolumn{2}{l}{}                                                                   & 1\% & 3\% & 5\% \\ \hline
\multirow{3}{*}{\textbf{TotalSeg}} & \multicolumn{1}{l|}{\textbf{DSC}}           & 92.98         & 90.72         & 88.83         \\
                                         & \multicolumn{1}{l|}{\textbf{\#(MB)}}        & 6.53           & 4.50           & 3.28           \\
                                         & \multicolumn{1}{l|}{\textbf{Rel \#}} & 0.43           & 0.30           & 0.22           \\ \hline
\multirow{3}{*}{\textbf{ChestOrgan}}  & \multicolumn{1}{l|}{\textbf{DSC}}           & 78.26         & 77.16         & 74.88         \\
                                         & \multicolumn{1}{l|}{\textbf{\#(MB)}}        & 3.39           & 2.85           & 1.23           \\
                                         & \multicolumn{1}{l|}{\textbf{Rel \#}} & 0.23           & 0.19           & 0.08           \\ \hline
\multirow{3}{*}{\textbf{HNOrgan}}     & \multicolumn{1}{l|}{\textbf{DSC}}           & 83.97         & 82.24         & 80.27         \\
                                         & \multicolumn{1}{l|}{\textbf{\#(MB)}} & 4.18           & 4.04           & 1.88           \\
                                         & \multicolumn{1}{l|}{\textbf{Rel \#}} & 0.28           & 0.27           & 0.12           \\ \hline
\multirow{3}{*}{\textbf{EsoOrgan}}   & \multicolumn{1}{l|}{\textbf{DSC}}           & 86.94         & 85.97         & --             \\
                                         & \multicolumn{1}{l|}{\textbf{\#(MB)}} & 0.67           & 0.57           & --             \\
                                         & \multicolumn{1}{l|}{\textbf{Rel \#}} & 0.04           & 0.04           & --             \\ \hline
\end{tabular}
}
\vspace{-1em}
\end{table}

\subsection{Ablation Study Results}

{\noindent \bf Effectiveness of General Encoder:} 
\textcolor{\highlight}{To demonstrate the importance of freezing the General Encoder when learning subsequent tasks,}
we compare the segmentation performance with and without freezing the General Encoder when continually learning on new datasets (using two CSS orders with two learning steps). Results are summarized in Table~\ref{tab:abl_unfrz}. First, it is observed that without freezing the General Encoder, the model has catastrophic forgetting, e.g., segmentation DSC of the old dataset in TotalSegmentator $\rightarrow$ ChestOrgan decreases from 93.24\% to 51.42\% as compared to that with the frozen encoder. Second, the performance for segmenting the new dataset is similar regardless of the encoder status (freezing or trainable). For instance, 84.14\% vs. 84.35\% DSC of HNOrgan dataset is achieved in TotalSegmentator $\rightarrow$ HNOrgan. The experimental results demonstrate that a well-trained and subsequently frozen General Encoder could generalize well to support specialized tasks.

{\noindent \bf Effectiveness of Decoder Pruning:}
Table~\ref{tab:trim} shows the detailed decoder pruning results. Several conclusions can be drawn. First, the proposed decoder pruning method achieves a good trade-off between model complexity and accuracy reduction. For example, for the TotalSegmentator decoder, with 1\% DSC decrease, the number of parameters is reduced from 15.07 MB to 6.53 MB with a relative 43\% of the original decoder size. As the larger performance drop is allowed, e.g., 3\% and 5\% DSC decrease, the size of pruned decoder decreases to 30\% and 22\% of the original decoder, respectively. Second, as the number of segmented organs becomes smaller, a higher compressed ratio can be achieved. With 1\% DSC performance decrease, the pruned ChestOrgan decoder (segmenting 31 organs) has 3.39 MB parameters as compared to 6.53 MB of pruned TotalSegmentator decoder.  Third, the EsoOrgan decoder has the highest model compression ratio with only 0.67 MB parameters (4\% of original decoder size). This indicates that domain-incremental segmentation may be an easier task as compared to class-incremental continual segmentation.


{\noindent \bf Effectiveness of Merging Module:}
Table~\ref{tab:merge} presents the segmentation results using two merging methods. It is observed that a simple ensemble-based merging method exhibits decreased performance in all metrics on all datasets. The proposed anomaly-aware output merging significantly boosts the performance on the EsoOrgan dataset (DSC: 80.22\% to 86.94\%, HD95: 7.62 to 5.04mm, ASD: 1.92 to 1.11mm). This demonstrates the effectiveness and importance of the abnormal detection module. The proposed merging module can identify the esophageal tumor and subsequently generate a high confidence score for the EsoOrgan decoder suitable for segmenting advanced esophageal cancer patients. In contrast, the ensemble method could not differentiate if there exists abnormality in an image. Hence, averaging the esophagus predictions from three decoders that predict the esophagus leads to significantly decreased performance. 


\begin{table}
\centering
\caption{Quantitative results of using different output merging methods. Mean DSC (\%), HD95 (mm) and ASD (mm) are evaluated. Better performance is indicated in bold. }
\label{tab:merge}
\vspace{2mm}
\scalebox{0.82}{
\begin{tabular}{lllllll}
\hline
\textbf{}                                & \multicolumn{3}{c}{\textbf{Ensemble}}                            & \multicolumn{3}{c}{\textbf{Anomaly-aware merging}}          \\ \cline{2-7} 
\multicolumn{1}{l}{}                    & \textbf{DSC} & \textbf{HD95} & \multicolumn{1}{l|}{\textbf{ASD}} & \textbf{DSC}   & \textbf{HD95} & \textbf{ASD}  \\ \hline
\multicolumn{1}{l|}{\textbf{TotalSeg}}   & 88.59        & 4.41          & \multicolumn{1}{l|}{1.09}         & \textbf{92.98} & \textbf{4.09} & \textbf{0.98} \\
\multicolumn{1}{l|}{\textbf{ChestOrgan}} & 76.78        & 9.44          & \multicolumn{1}{l|}{1.89}         & \textbf{78.26} & \textbf{9.17} & \textbf{1.82} \\
\multicolumn{1}{l|}{\textbf{HNOrgan}}    & 77.84        & 2.65          & \multicolumn{1}{l|}{0.67}         & \textbf{83.97} & \textbf{2.22} & \textbf{0.59} \\
\multicolumn{1}{l|}{\textbf{EsoOrgan}}   & 80.22        & 7.62          & \multicolumn{1}{l|}{1.92}         & \textbf{86.94} & \textbf{5.04} & \textbf{1.11} \\ \hline
\end{tabular}
}
\vspace{-1em}
\end{table}

{\noindent \bf Alternative Training Dataset for the General Encoder:}
\textcolor{\edited}{We recommend starting with TotalSegentator as $D_1$ to train the General Encoder as it covers most body parts with a large set of labeled organs for comprehensive feature extraction. However, this is not a hard requirement. Alternatively, other datasets can also be used as the starting
dataset to train the General Encoder. 
When training the General Encoder using the ChestOrgan dataset with much less training scans (292 vs. 1204 CT scans) and organ classes (31 vs. 103 anatomical structures), a tolerable performance drop ($<$1\% Dice) of our method is observed in the final results of CSS Order A. The assumed reason is that CT scans in the ChestOrgan dataset covers most of the torso region with diverse anatomies, which allows the General Encoder to learn sufficient representative features. Hence, General Encoder trained with ChestOrgan exhibits similar performance as the one trained using TotalSegmentator. 
}


\section{Conclusion}
In this work, we propose a new CSS framework to continually segment a total of 143 whole-body organs from four partially labeled datasets. With the \textcolor{\edited}{trained and frozen} General Encoder and continually-added and architecturally optimized decoders, our model avoids catastrophic forgetting while effectively segmenting new organs with high accuracy. We further propose a body-part and anomaly-aware output merging module to combine organ predictions originating from different decoders and incorporate both healthy and pathological organs appearing in different datasets.  
Continually trained and validated on 3D CT scans of 2500+ patients of four datasets, our single network can segment 143 whole-body organs with very high accuracy, closely reaching the upper bound performance level by training four separate segmentation models.

{\small
\bibliographystyle{ieee_fullname}
\bibliography{egbib, ref_ciss}
}

\newpage
\section*{}
\newpage
\section*{Appendix}
\appendix
\counterwithin{figure}{section}
\counterwithin{table}{section}


\section{Dataset Details}
\label{supp:dataset}

We describe the dataset details (one public and three private multi-organ datasets) used in our experiment here. For the public dataset TotalSegmentator~\cite{wasserthal2022totalsegmentator}, it consists of 1204 CT scans of different body parts with total 103 labeled anatomical structures (26 major organs, 59 bone instances, 10 muscles, and 8 vessels). Note that the face label is removed as it is an artificially created label for patient de-identification purpose after blurring the facial area. The detailed organ 103 organ instance list can be found in the link \url{https://github.com/wasserth/TotalSegmentator}. For the three in-house multi-organ datasets, they are head \& neck organ dataset (denoted as HNOrgan), chest organ dataset (denoted as ChestOrgan) and dedicated esophageal cancer dataset (denoted as EsoOrgan). In HNOrgan, each of the 447 head and neck CT scans has 13 head and neck organs labeled: brainstem, eye (left and right), lens (left and right), optic nerve (left and right), optic chiasm,  parotid (left and right), spinal cord, temporomandibular joint (TMJoint, left and right).  ChestOrgan contains 292 chest CT scans with 31 chest anatomical structures annotated including major organs, muscles, arteries and veins. The detailed list is as follow: esophagus, sternum, thyroid left, thyroid right, trachea, bronchus left, bronchus right, anterior cervical muscle, scalenus muscle, scalenus anterior muscle, sternocleidomastoid muscle, ascending aorta, descending aorta, aorta arch, common carotid artery left, common carotid artery right, pulmonary artery, subclavian artery left, subclavian artery right, vertebral artery left, vertebral artery right, azygos vein, brachiocephalic vein left, brachiocephalic vein right, internal jugular vein left, internal jugular vein right, pulmonary vein, subclavian vein left, subclavian vein right, inferior vena cava, superior vena cava. There are four organs in ChestOrgan that are overlapped with organs in TotalSegmentator (esophagus, pulmonary artery, superior vena cava, trachea). The EsoOrgan collects 640 diagnostic CT scans of advanced esophageal cancer patient where only the esophagus is labeled. By combining all datasets, we have total 103+27+13 organ classes from 2583 unique patients.  For each of these four datasets, $20\%$ are randomly chosen as the testing set, while the rest is used as training + validation.  Detailed training/validation dataset split in the decoder optimization module can be found in the Implementation Details section of this supplementary material.

In addition, for the purpose of training and validating our abnormality detection module, we further collect CT scans from 304 esophageal (private) and 625 lung cancer (public with tumor labels) patients where the 3D tumor masks are delineated at the pixel level. These combine as the lung/esophageal tumor classes from additional 929 patients.


\section{Implementation Details}
\label{supp:implement}


The default nnUNet backbone in 3D full resolution setting is adopted in our work, including a 6-block encoding path and a 5-block decoding path. Each encoding block consists of the following consecutive operations: a convolution, an instance normalization, a Leaky ReLU unit, followed by max-pooling operator. 


The total training epoch for the baseline TotalSegmentator is 8000 with 250 iterations per epoch, and the training epoch for each of the in-house datasets (served as performance upper bound) is 1000 with 250 iterations per epoch. The batch size is 2. The optimizer is stochastic gradient descent with a Polynomial learning rate policy. The initial learning rate is 0.01 with a Nesterov momentum of 0.99. Default ``moreDA'' data augmentation is adopted, e.g., horizontal flipping, random rotation in the x-y plane with $\pm 10$ degrees, intensity scaling with a ratio between $[0.75, 1.25]$, adding Gaussian noise with zero mean and $[0, 0.1]$ variance. The total average training time is 2.5 GPU days per 1000 epochs. For model inference, the average running time for the proposed framework, before the decoding path optimization, is approximately 15 minutes per patient. After the decoding path optimization, the average inference time is less than 5 minutes per patient. All models are developed using PyTorch and trained on one NVIDIA A100 GPU. 

\begin{table*}[]
\centering
\caption{The detailed auto-searched and pruned decoding architecture based on nnUNet.
Note that decoder block 5 refers to the deepest decoding block. \label{tab:decoder_opt}}
\scalebox{0.83}{
\begin{tabular}{llccccc}
                                                          &        & \textbf{Decoder Block 5} & \textbf{Decoder Block 4} & \textbf{Decoder Block 3} & \textbf{Decoder Block 2} & \textbf{Decoder Block 1} \\ \hline
\multicolumn{1}{l|}{\multirow{2}{*}{\textbf{TotalSeg}}}   & NAS    & P3D                      & P3D                      & 2D                       & P3D                      & 2D                       \\
\multicolumn{1}{l|}{}                                     & Pruned & P3D                      & P3D                      & 2D                       & Projection               & 2D                       \\ \hline
\multicolumn{1}{l|}{\multirow{2}{*}{\textbf{ChestOrgan}}} & NAS    & 3D                       & P3D                      & P3D                      & 2D                       & 2D                       \\
\multicolumn{1}{l|}{}                                     & Pruned & Projection               & P3D                      & P3D                      & 2D                       & 2D                       \\ \hline
\multicolumn{1}{l|}{\multirow{2}{*}{\textbf{HNOrgan}}}    & NAS    & 3D                       & P3D                      & 3D                       & P3D                      & P3D                      \\
\multicolumn{1}{l|}{}                                     & Pruned & Projection               & P3d                      & 3D                       & P3D                      & P3D                      \\ \hline
\multicolumn{1}{l|}{\multirow{2}{*}{\textbf{EsoOrgan}}}   & NAS    & P3D                      & 2D                       & Projection               & 2D                       & Projection               \\
\multicolumn{1}{l|}{}                                     & Pruned & Projection               & Projection               & Projection               & Projection               & Projection               \\ \hline
\end{tabular}%
}
\end{table*}


\noindent\textbf{Decoder Optimization -- NAS Setting.}
For NAS, we divide each dataset into 1) $60\%$ for network training, 2) $30\%$ for NAS training, and 3) $10\%$ for validation and ablation evaluation. To train the learnable weight for selecting the architecture of each decoding block, we first fix the convolution kernel selection weight to $\frac{1}{C_t}$ for $40\%$ of the total epochs, where $C_t$ denotes the number of classes of the $t^{th}$ task. Then we alternatively update the convolution kernel selection weight and the decoder parameters for the rest epochs. The initial learning rate is set to 0.01 for all tasks. The learning rate is decayed following the Polynomial learning rate policy. After NAS training is complete, we follow the same `moreDA' data augmentation scheme and retrain the searched decoding path from scratch using the re-divided dataset in a 4:1 `training-validation' ratio. The searched decoding blocks for each task are shown in Table~\ref{tab:decoder_opt}. 

\noindent\textbf{Decoder Optimization -- Pruning Setting.}
We perform a block-wise teacher-student \ac{kd}, aiming to further compress the decoder by replacing the searched convolutional kernels with the projection kernels. The mean-square error loss is used to match the feature map outputs of the teacher block to the student block. To ease the optimization difficulty, we first distill the deeper decoding blocks (lower resolution), then move to shallower blocks. Once the \ac{kd} training of the deeper block is saturated, we freeze the deeper student blocks and move to the shallower ones. 
When conducting this block-wise KD, the shallower convolutional block needs to choose if to receive output feature maps from the deeper teacher blocks or those from the student block. Under this setting, if the feature map difference from previous teacher and student block is large, it would affect the subsequent feature response in the next shallower block causing the degenerated segmentation. To conquer this, we employ a simple yet effective approach: Once the distillation of the student block is finished, we use a smaller learning rate (e.g., $0.1\times$) and finetune the shallower teacher block using the deeper student block's output feature maps. We monitor the before-and-after performance drop. If the drop is less than $1\%$ in terms of Dice score, we keep the deeper student block, otherwise, we restore the teacher block. To speed up the training process, block-wise side supervision is also used during training. The pruned decoder for each task is demonstrated in Table~\ref{tab:decoder_opt}.

\noindent\textbf{Body-part and Anomaly-aware Decoder Merging.}
To generate the final 143-class organ segmentation output, we need to combine and merge the predictions from all decoders. There are two major issues in this step. First, since each training dataset/task often covers a specific body part, the task-specific decoder might generate false positives in body parts that are not covered in this specific dataset (because that decoder never sees other anatomic regions). We propose a straightforward yet effective solution to reduce these false positives due to the body part coverage: for each decoder, we pre-compute the body part coverage rate using all the data in this dataset/task. In this way, for a specific decoder, voxels outside the covered body parts would have a lower/zero weight, and we can use this weight to significantly decrease the confidence score of the false positive segmentations out of the covered body parts. Specifically, we first generate the body part map using an automated body part regression algorithm~\cite{yan2018unsupervised}. Then, by overlapping the bounding box of all labeled organs to the body part map, we compute the volume-wise overlapping ratio between the bounding box and the body part map. Then, a body part distribution map is generated for a patient, e.g., 80\% in the chest, and 20\% in the abdomen. This calculation is repeated for all patients in the dataset, and finally averaged to get a pre-computed body part distribution map $\hat{Y}^t_{\beta}$. The detailed body part distributions for the three in-house datasets are illustrated in Table~\ref{tab:bodyparts}.

\begin{table}[]
\centering
\caption{The body part distributions of the in-house datasets.
\label{tab:bodyparts}}
\scalebox{0.90}{
\begin{tabular}{lllll}
                                & Head & Chest  & Abdomen & Pelvic \\ \hline
\multicolumn{1}{l|}{ChestOrgan} & 8.2\%        & 89.4\% & 2.4\%   & 0\%          \\
\multicolumn{1}{l|}{HNOrgan}    & 96.5\%       & 3.5\%  & 0\%     & 0\%          \\
\multicolumn{1}{l|}{EsoOrgan}   & 0\%          & 96.3\% & 3.7\%   & 0\%          \\ \hline
\end{tabular}%
}
\end{table}

The second issue is that some decoders might not see the patients with abnormalities (e.g., tumors). Hence, the predictions may have false segmentation in the anomaly region. To resolve this problem, we first supplement the framework with an anomaly segmentation head. In our work, we use the esophageal and lung tumor dataset to train this head as an illustration. More abnormality datasets can be utilized, such as DeepLesion~\cite{yan2018deeplesion}. Then, we exploit the tumor predictions $\hat{Y}^{\epsilon}$ to generate an anomaly weighting map. The averaged tumor size $p^{\epsilon}$ is pre-calculated using the annotated tumor mask and used as the standard deviation of a Gaussian filter of zero-mean to further smooth the tumor prediction $\hat{Y}^{\epsilon}$. Here, we assume that, at location $j$, the prediction $\hat{Y}^t(j)$ of the $t^{th}$ decoder is less confident if $\hat{Y}^{\epsilon}(j)$ is of high value. 

To combine the predictions, we perform a voxel-wise selection by choosing the most confident prediction from all decoding heads, considering both the body part distribution map and the anomaly map. As the entropy function produces the highest value when the input closes to 0.5 (the most uncertain prediction), we could find the most confident prediction when the input closes to 0 or 1. Eq.~(3) in the main text is used to combine the body part map $\hat{Y}^t_{\beta}$ and anomaly distribution map $\hat{Y}^{\epsilon}$. When there is no tumor prediction $\hat{Y}^{\epsilon}(j)=0$ and the organ prediction is within the decoder's body part distribution range $\hat{Y}^{t}_{\beta}(j)=1$, the output score is considered as confident and sets $M^t(j) =0$. On the other hand, the other states are considered uncertain and set $M^t(j)=0.5$. The confidence map is generated using Eq.~(4) of the main text. For a voxel at location $j$,  using Eq.~(5), we collect the confidence values from all tasks whose foreground prediction is not in the background. The final output class $\hat{\mathbf{Y}}(j)$ is determined using the prediction $\hat{Y}^t(j)$, of which with the smallest $H^t(j)$.



\section{Additional Ablation Results and Analysis}
\label{supp:ablation}
\begin{table*}
\centering
\caption{Mean DSC (\%, $\uparrow$) and forgetting rate (\%, $\downarrow$) of our method and other comparison methods in each dataset/step of two continual segmentation orders. The last column `All Learned Classes' lists the mean DSC and average forgetting rate~\cite{chaudhry2018riemannian} over all learned organs/classes at each step. The DSC in TotalSegmentator at step 1 of all comparison methods is the upper bound 93.24\%, while DSC for our method is slightly lower with 92.98\% at step 1 due to the decoder compression/pruning. (Sub-figures in Fig. \ref{M-fig:step_all} from left to right are corresponding to the numeric results in column `All Learned Classes', `TotalSeg', `ChestOrgan' and `HNOrgan' of this table.)}
\label{tab:step_task}
\vspace{1em}
\scalebox{0.83}{
\begin{tabular}{@{}lrrrrrrrllrr@{}}
\toprule
                                                                               & \multicolumn{1}{c|}{}                                & \multicolumn{2}{c|}{\textbf{TotalSeg (103)}}                            & \multicolumn{2}{c|}{\textbf{ChestOrgan (31)}}                                                  & \multicolumn{2}{c|}{\textbf{HNOrgan (13)}}                                                     & \multicolumn{2}{c|}{\textbf{EsoOrgan (1)}}                                & \multicolumn{2}{c}{\textbf{All Learned Classes}}                            \\ \cmidrule(l){3-12} 
\multirow{-2}{*}{\textbf{Methods}}                                             & \multicolumn{1}{c|}{\multirow{-2}{*}{\textbf{Step}}} & \multicolumn{1}{l}{\textbf{DSC}} & \multicolumn{1}{l|}{\textbf{Forget}} & \multicolumn{1}{l}{\textbf{DSC}}             & \multicolumn{1}{l|}{\textbf{Forget}}          & \multicolumn{1}{l}{\textbf{DSC}}             & \multicolumn{1}{l|}{\textbf{Forget}}          & \textbf{DSC}              & \multicolumn{1}{l|}{\textbf{Forget}}          & \multicolumn{1}{l}{\textbf{DSC}} & \multicolumn{1}{l}{\textbf{Avg. Forget}} \\ \midrule
\textbf{upper bound}                                                           & \multicolumn{1}{l|}{}                                & 93.24                            & \multicolumn{1}{r|}{---}             & 78.45                                        & \multicolumn{1}{r|}{---}                      & 84.35                                        & \multicolumn{1}{r|}{---}                      & \multicolumn{1}{r}{87.15} & \multicolumn{1}{r|}{---}                      & 89.02                            & ---                                      \\ \midrule\midrule
                                                                               & \multicolumn{11}{c}{\textbf{Order A: TotalSeg $\rightarrow$ ChestOrgan $\rightarrow$ HNOrgan $\rightarrow$ EsoOrgan}}                                                                                                                                                                                                                                                                                                                                                                  \\ \midrule
                                                                               & \multicolumn{1}{r|}{2}                               & 45.80                            & \multicolumn{1}{r|}{50.87}           & 78.40                                        & \multicolumn{1}{r|}{---}                      & \multicolumn{1}{l}{\cellcolor[HTML]{D9D9D9}} & \multicolumn{1}{l|}{\cellcolor[HTML]{D9D9D9}} & \cellcolor[HTML]{D9D9D9}  & \multicolumn{1}{l|}{\cellcolor[HTML]{D9D9D9}} & 50.56                            & 50.87                                    \\
                                                                               & \multicolumn{1}{r|}{3}                               & 11.68                            & \multicolumn{1}{r|}{87.48}           & 25.66                                        & \multicolumn{1}{r|}{67.27}                    & 84.22                                        & \multicolumn{1}{r|}{---}                      & \cellcolor[HTML]{D9D9D9}  & \multicolumn{1}{l|}{\cellcolor[HTML]{D9D9D9}} & 20.86                            & 77.37                                    \\
\multirow{-3}{*}{\textbf{MiB~\cite{cermelli2020modeling}}}                     & \multicolumn{1}{r|}{4}                               & 7.65                             & \multicolumn{1}{r|}{91.80}           & 19.24                                        & \multicolumn{1}{r|}{75.46}                    & 6.37                                         & \multicolumn{1}{r|}{92.43}                    & \multicolumn{1}{r}{86.92} & \multicolumn{1}{r|}{---}                      & 8.51                             & 86.56                                    \\ \midrule
                                                                               & \multicolumn{1}{r|}{2}                               & 48.50                            & \multicolumn{1}{r|}{47.98}           & 77.78                                        & \multicolumn{1}{r|}{---}                      & \multicolumn{1}{l}{\cellcolor[HTML]{D9D9D9}} & \multicolumn{1}{l|}{\cellcolor[HTML]{D9D9D9}} & \cellcolor[HTML]{D9D9D9}  & \multicolumn{1}{l|}{\cellcolor[HTML]{D9D9D9}} & 54.40                            & 47.98                                    \\
                                                                               & \multicolumn{1}{r|}{3}                               & 13.68                            & \multicolumn{1}{r|}{85.33}           & 28.74                                        & \multicolumn{1}{r|}{63.04}                    & 84.21                                        & \multicolumn{1}{r|}{---}                      & \cellcolor[HTML]{D9D9D9}  & \multicolumn{1}{l|}{\cellcolor[HTML]{D9D9D9}} & 23.08                            & 74.19                                    \\
\multirow{-3}{*}{\textbf{$\text{ILT}^\dagger$~\cite{michieli2019incremental}}} & \multicolumn{1}{r|}{4}                               & 10.87                            & \multicolumn{1}{r|}{88.34}           & 27.87                                        & \multicolumn{1}{r|}{64.17}                    & 6.39                                         & \multicolumn{1}{r|}{92.42}                    & \multicolumn{1}{r}{85.75} & \multicolumn{1}{r|}{---}                      & 11.99                            & 81.64                                    \\ \midrule
                                                                               & \multicolumn{1}{r|}{2}                               & 59.13                            & \multicolumn{1}{r|}{36.59}           & 76.52                                        & \multicolumn{1}{r|}{---}                      & \multicolumn{1}{l}{\cellcolor[HTML]{D9D9D9}} & \multicolumn{1}{l|}{\cellcolor[HTML]{D9D9D9}} & \cellcolor[HTML]{D9D9D9}  & \multicolumn{1}{l|}{\cellcolor[HTML]{D9D9D9}} & 62.40                            & 36.59                                    \\
                                                                               & \multicolumn{1}{r|}{3}                               & 39.46                            & \multicolumn{1}{r|}{57.68}           & 49.19                                        & \multicolumn{1}{r|}{35.72}                    & 83.17                                        & \multicolumn{1}{r|}{---}                      & \cellcolor[HTML]{D9D9D9}  & \multicolumn{1}{l|}{\cellcolor[HTML]{D9D9D9}} & 45.10                            & 46.70                                    \\
\multirow{-3}{*}{\textbf{PLOP~\cite{douillard2021plop}}}                       & \multicolumn{1}{r|}{4}                               & 37.30                            & \multicolumn{1}{r|}{59.99}           & 51.74                                        & \multicolumn{1}{r|}{32.38}                    & 25.38                                        & \multicolumn{1}{r|}{69.48}                    & \multicolumn{1}{r}{82.90} & \multicolumn{1}{r|}{---}                      & 39.01                            & 53.95                                    \\ \midrule
                                                                               & \multicolumn{1}{r|}{2}                               & 52.57                            & \multicolumn{1}{r|}{43.62}           & 78.48                                        & \multicolumn{1}{r|}{---}                      & \multicolumn{1}{l}{\cellcolor[HTML]{D9D9D9}} & \multicolumn{1}{l|}{\cellcolor[HTML]{D9D9D9}} & \cellcolor[HTML]{D9D9D9}  & \multicolumn{1}{l|}{\cellcolor[HTML]{D9D9D9}} & 57.74                            & 43.62                                    \\
                                                                               & \multicolumn{1}{r|}{3}                               & 13.59                            & \multicolumn{1}{r|}{85.42}           & 29.05                                        & \multicolumn{1}{r|}{62.99}                    & 84.36                                        & \multicolumn{1}{r|}{---}                      & \cellcolor[HTML]{D9D9D9}  & \multicolumn{1}{l|}{\cellcolor[HTML]{D9D9D9}} & 22.86                            & 74.21                                    \\
\multirow{-3}{*}{\textbf{LISMO~\cite{liu2022learning}}}                        & \multicolumn{1}{r|}{4}                               & 10.82                            & \multicolumn{1}{r|}{88.40}           & 28.24                                        & \multicolumn{1}{r|}{64.02}                    & 6.30                                         & \multicolumn{1}{r|}{92.54}                    & \multicolumn{1}{r}{87.12} & \multicolumn{1}{r|}{---}                      & 12.11                            & 81.65                                    \\ \midrule
                                                                               & \multicolumn{1}{r|}{2}                               & 92.98                            & \multicolumn{1}{r|}{0.00}            & 78.26                                        & \multicolumn{1}{r|}{---}                      & \multicolumn{1}{l}{\cellcolor[HTML]{D9D9D9}} & \multicolumn{1}{l|}{\cellcolor[HTML]{D9D9D9}} & \cellcolor[HTML]{D9D9D9}  & \multicolumn{1}{l|}{\cellcolor[HTML]{D9D9D9}} & 88.27                            & 0.00                                     \\
                                                                               & \multicolumn{1}{r|}{3}                               & 92.98                            & \multicolumn{1}{r|}{0.00}            & 78.26                                        & \multicolumn{1}{r|}{0.00}                     & 83.97                                        & \multicolumn{1}{r|}{---}                      & \cellcolor[HTML]{D9D9D9}  & \multicolumn{1}{l|}{\cellcolor[HTML]{D9D9D9}} & 87.88                            & 0.00                                     \\
\multirow{-3}{*}{\textbf{Ours}}                                                & \multicolumn{1}{r|}{4}                               & 92.98                            & \multicolumn{1}{r|}{0.00}            & 78.26                                        & \multicolumn{1}{r|}{0.00}                     & 83.97                                        & \multicolumn{1}{r|}{0.00}                     & \multicolumn{1}{r}{86.94} & \multicolumn{1}{r|}{---}                      & 87.87                            & 0.00                                     \\ \midrule\midrule
                                                                               & \multicolumn{11}{c}{\textbf{Order B: TotalSeg $\rightarrow$ HNOrgan $\rightarrow$ ChestOrgan $\rightarrow$ EsoOrgan}}                                                                                                                                                                                                                                                                                                                                                                  \\ \midrule
                                                                               & \multicolumn{1}{r|}{2}                               & 21.96                            & \multicolumn{1}{r|}{76.45}           & \multicolumn{1}{l}{\cellcolor[HTML]{D9D9D9}} & \multicolumn{1}{l|}{\cellcolor[HTML]{D9D9D9}} & 84.49                                        & \multicolumn{1}{r|}{---}                      & \cellcolor[HTML]{D9D9D9}  & \multicolumn{1}{l|}{\cellcolor[HTML]{D9D9D9}} & 29.42                            & 76.45                                    \\
                                                                               & \multicolumn{1}{r|}{3}                               & 10.72                            & \multicolumn{1}{r|}{88.50}           & 78.46                                        & \multicolumn{1}{r|}{---}                      & 6.38                                         & \multicolumn{1}{r|}{92.45}                    & \cellcolor[HTML]{D9D9D9}  & \multicolumn{1}{l|}{\cellcolor[HTML]{D9D9D9}} & 23.85                            & 90.48                                    \\
\multirow{-3}{*}{\textbf{MiB~\cite{cermelli2020modeling}}}                     & \multicolumn{1}{r|}{4}                               & 10.35                            & \multicolumn{1}{r|}{88.90}           & 65.63                                        & \multicolumn{1}{r|}{16.35}                    & 6.29                                         & \multicolumn{1}{r|}{92.55}                    & \multicolumn{1}{r}{86.79} & \multicolumn{1}{r|}{---}                      & 20.20                            & 65.94                                    \\ \midrule
                                                                               & \multicolumn{1}{r|}{2}                               & 21.62                            & \multicolumn{1}{r|}{76.81}           & \multicolumn{1}{l}{\cellcolor[HTML]{D9D9D9}} & \multicolumn{1}{l|}{\cellcolor[HTML]{D9D9D9}} & 84.25                                        & \multicolumn{1}{r|}{---}                      & \cellcolor[HTML]{D9D9D9}  & \multicolumn{1}{l|}{\cellcolor[HTML]{D9D9D9}} & 29.09                            & 76.81                                    \\
                                                                               & \multicolumn{1}{r|}{3}                               & 14.10                            & \multicolumn{1}{r|}{84.88}           & 78.02                                        & \multicolumn{1}{r|}{---}                      & 8.48                                         & \multicolumn{1}{r|}{89.93}                    & \cellcolor[HTML]{D9D9D9}  & \multicolumn{1}{l|}{\cellcolor[HTML]{D9D9D9}} & 26.13                            & 87.41                                    \\
\multirow{-3}{*}{\textbf{$\text{ILT}^\dagger$~\cite{michieli2019incremental}}} & \multicolumn{1}{r|}{4}                               & 13.12                            & \multicolumn{1}{r|}{85.93}           & 67.28                                        & \multicolumn{1}{r|}{13.76}                    & 6.18                                         & \multicolumn{1}{r|}{92.66}                    & \multicolumn{1}{r}{85.52} & \multicolumn{1}{r|}{---}                      & 22.44                            & 64.12                                    \\ \midrule
                                                                               & \multicolumn{1}{r|}{2}                               & 45.11                            & \multicolumn{1}{r|}{51.62}           & \multicolumn{1}{l}{\cellcolor[HTML]{D9D9D9}} & \multicolumn{1}{l|}{\cellcolor[HTML]{D9D9D9}} & 83.59                                        & \multicolumn{1}{r|}{---}                      & \cellcolor[HTML]{D9D9D9}  & \multicolumn{1}{l|}{\cellcolor[HTML]{D9D9D9}} & 49.70                            & 50.87                                    \\
                                                                               & \multicolumn{1}{r|}{3}                               & 31.90                            & \multicolumn{1}{r|}{65.79}           & 76.13                                        & \multicolumn{1}{r|}{---}                      & 17.56                                        & \multicolumn{1}{r|}{78.99}                    & \cellcolor[HTML]{D9D9D9}  & \multicolumn{1}{l|}{\cellcolor[HTML]{D9D9D9}} & 38.99                            & 72.39                                    \\
\multirow{-3}{*}{\textbf{PLOP~\cite{douillard2021plop}}}                       & \multicolumn{1}{r|}{4}                               & 30.82                            & \multicolumn{1}{r|}{66.95}           & 70.18                                        & \multicolumn{1}{r|}{7.81}                     & 15.77                                        & \multicolumn{1}{r|}{81.13}                    & \multicolumn{1}{r}{83.41} & \multicolumn{1}{r|}{---}                      & 36.63                            & 51.96                                    \\ \midrule
                                                                               & \multicolumn{1}{r|}{2}                               & 24.36                            & \multicolumn{1}{r|}{73.88}           & \multicolumn{1}{l}{\cellcolor[HTML]{D9D9D9}} & \multicolumn{1}{l|}{\cellcolor[HTML]{D9D9D9}} & 84.35                                        & \multicolumn{1}{r|}{---}                      & \cellcolor[HTML]{D9D9D9}  & \multicolumn{1}{l|}{\cellcolor[HTML]{D9D9D9}} & 31.51                            & 73.88                                    \\
                                                                               & \multicolumn{1}{r|}{3}                               & 15.08                            & \multicolumn{1}{r|}{83.83}           & 78.47                                        & \multicolumn{1}{r|}{---}                      & 7.85                                         & \multicolumn{1}{r|}{90.69}                    & \cellcolor[HTML]{D9D9D9}  & \multicolumn{1}{l|}{\cellcolor[HTML]{D9D9D9}} & 26.84                            & 87.26                                    \\
\multirow{-3}{*}{\textbf{LISMO~\cite{liu2022learning}}}                        & \multicolumn{1}{r|}{4}                               & 14.04                            & \multicolumn{1}{r|}{84.94}           & 67.19                                        & \multicolumn{1}{r|}{14.37}                    & 6.15                                         & \multicolumn{1}{r|}{92.71}                    & \multicolumn{1}{r}{86.87} & \multicolumn{1}{r|}{---}                      & 23.09                            & 64.01                                    \\ \midrule
                                                                               & \multicolumn{1}{r|}{2}                               & 92.98                            & \multicolumn{1}{r|}{0.00}            & \multicolumn{1}{l}{\cellcolor[HTML]{D9D9D9}} & \multicolumn{1}{l|}{\cellcolor[HTML]{D9D9D9}} & 83.97                                        & \multicolumn{1}{r|}{---}                      & \cellcolor[HTML]{D9D9D9}  & \multicolumn{1}{l|}{\cellcolor[HTML]{D9D9D9}} & 91.27                            & 0.00                                     \\
                                                                               & \multicolumn{1}{r|}{3}                               & 92.98                            & \multicolumn{1}{r|}{0.00}            & 78.26                                        & \multicolumn{1}{r|}{---}                      & 83.97                                        & \multicolumn{1}{r|}{0.00}                     & \cellcolor[HTML]{D9D9D9}  & \multicolumn{1}{l|}{\cellcolor[HTML]{D9D9D9}} & 87.88                            & 0.00                                     \\
\multirow{-3}{*}{\textbf{Ours}}                                                & \multicolumn{1}{r|}{4}                               & 92.98                            & \multicolumn{1}{r|}{0.00}            & 78.26                                        & \multicolumn{1}{r|}{0.00}                     & 83.97                                        & \multicolumn{1}{r|}{0.00}                     & \multicolumn{1}{r}{86.94} & \multicolumn{1}{r|}{---}                      & 87.87                            & 0.00                                     \\ \bottomrule
\end{tabular}%
}
\end{table*}

\begin{figure*}
    \centering
    \includegraphics[width=0.40\paperwidth]{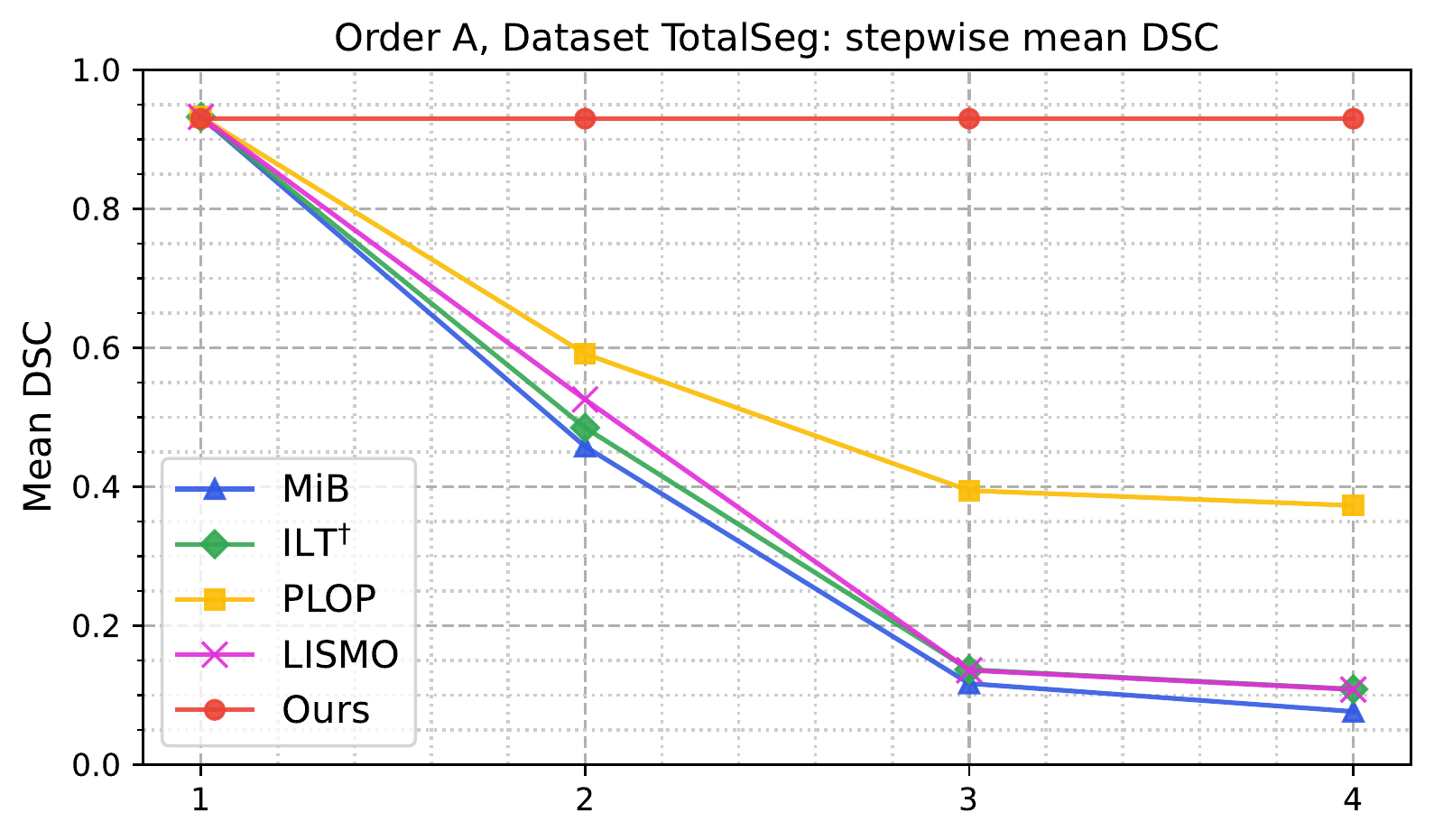}
    \includegraphics[width=0.40\paperwidth]{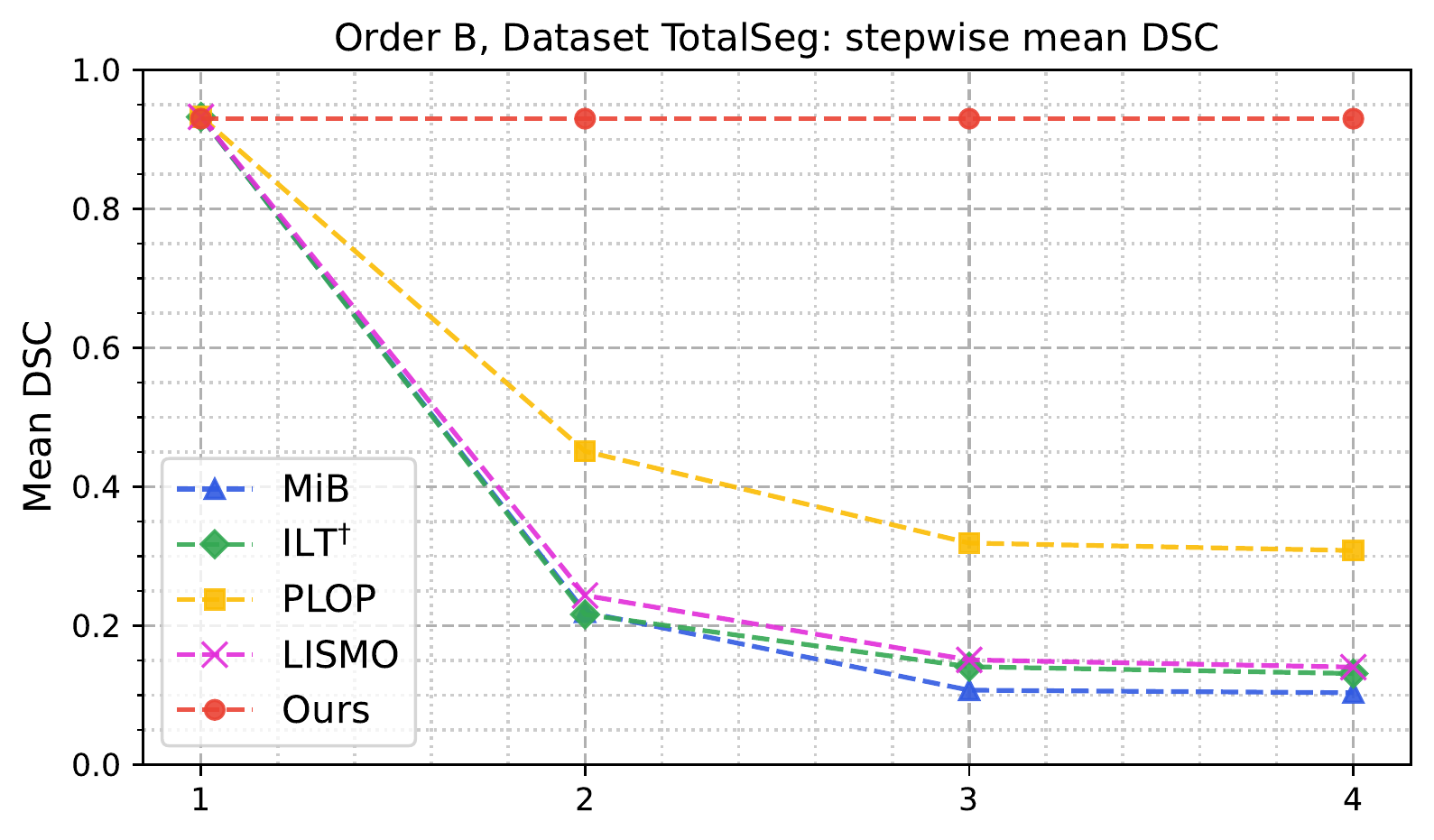}\\
    \includegraphics[trim={0.5em 0 0.5em 0}, clip, width=0.233\paperwidth]{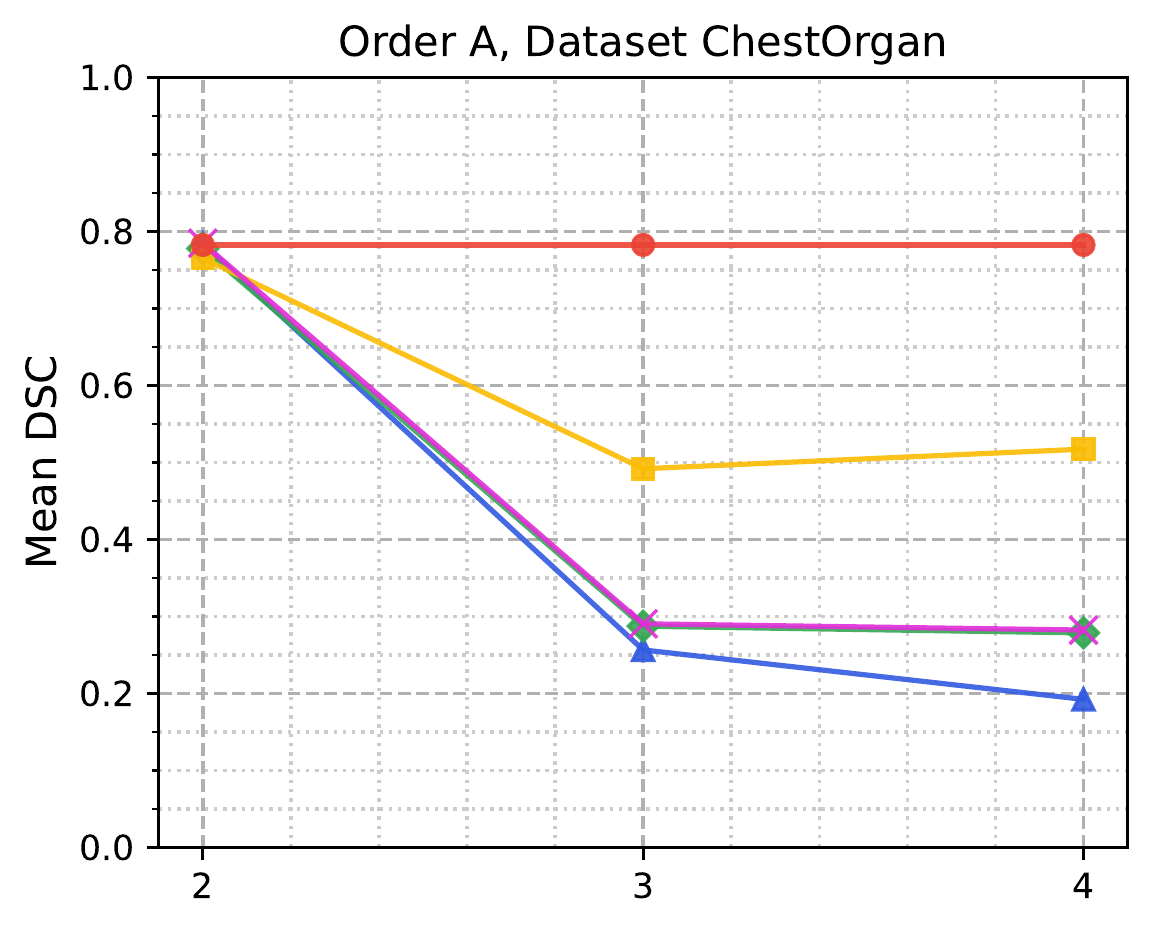}
    \includegraphics[trim={0.5em 0 0.5em 0}, clip, width=0.155\paperwidth]{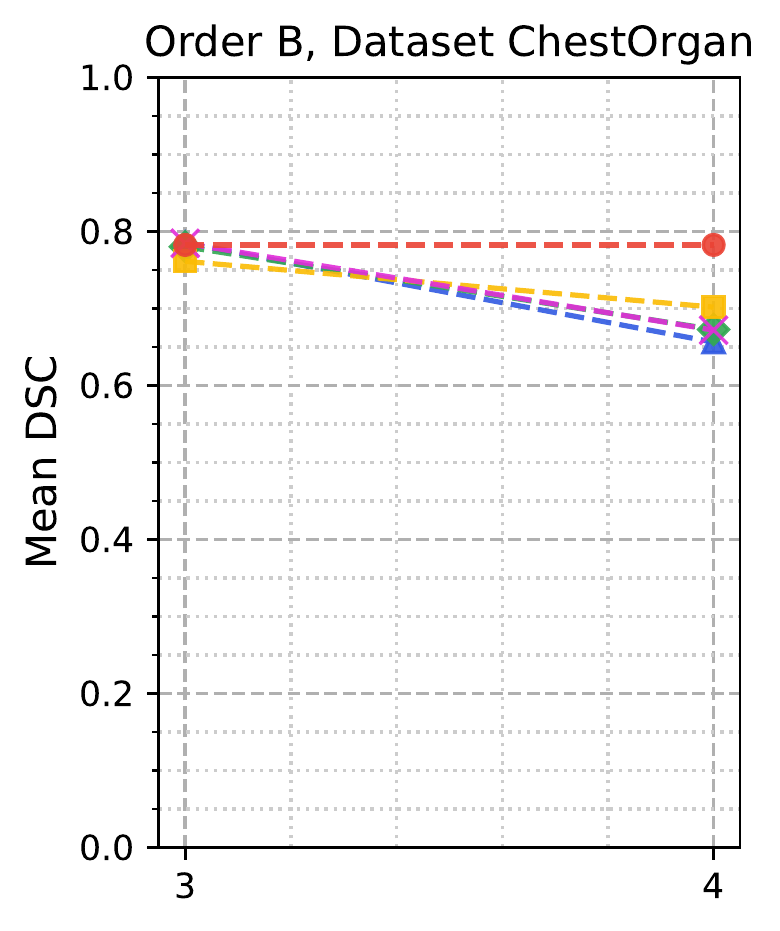}
    \includegraphics[trim={0.5em 0 0.5em 0}, clip, width=0.152\paperwidth]{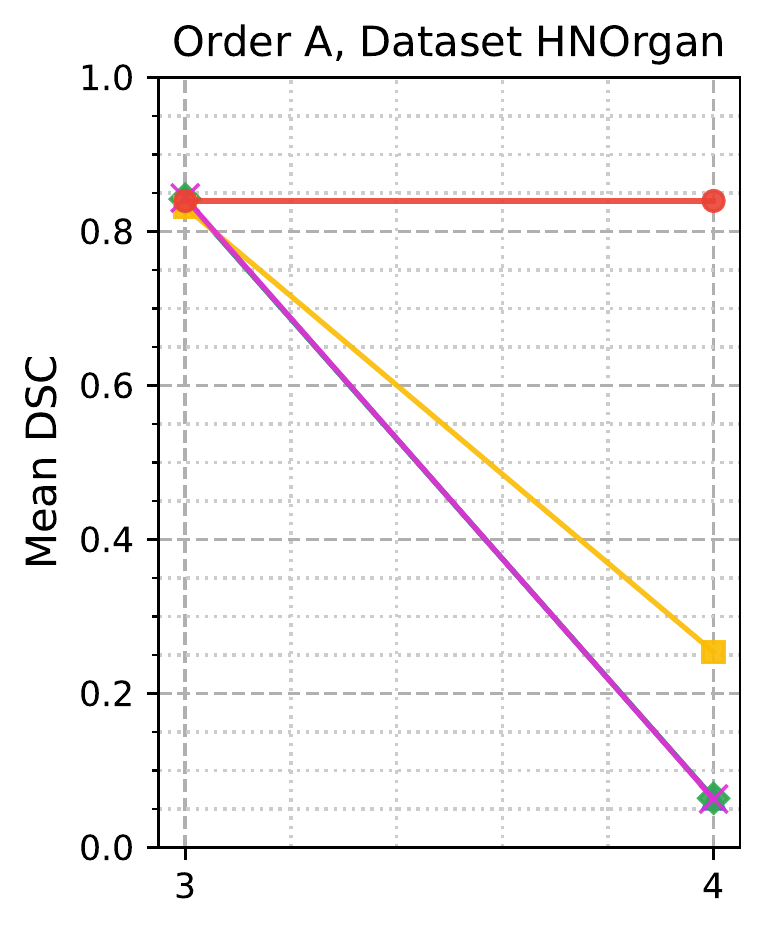}
    \includegraphics[trim={0.5em 0 0.5em 0}, clip, width=0.233\paperwidth]{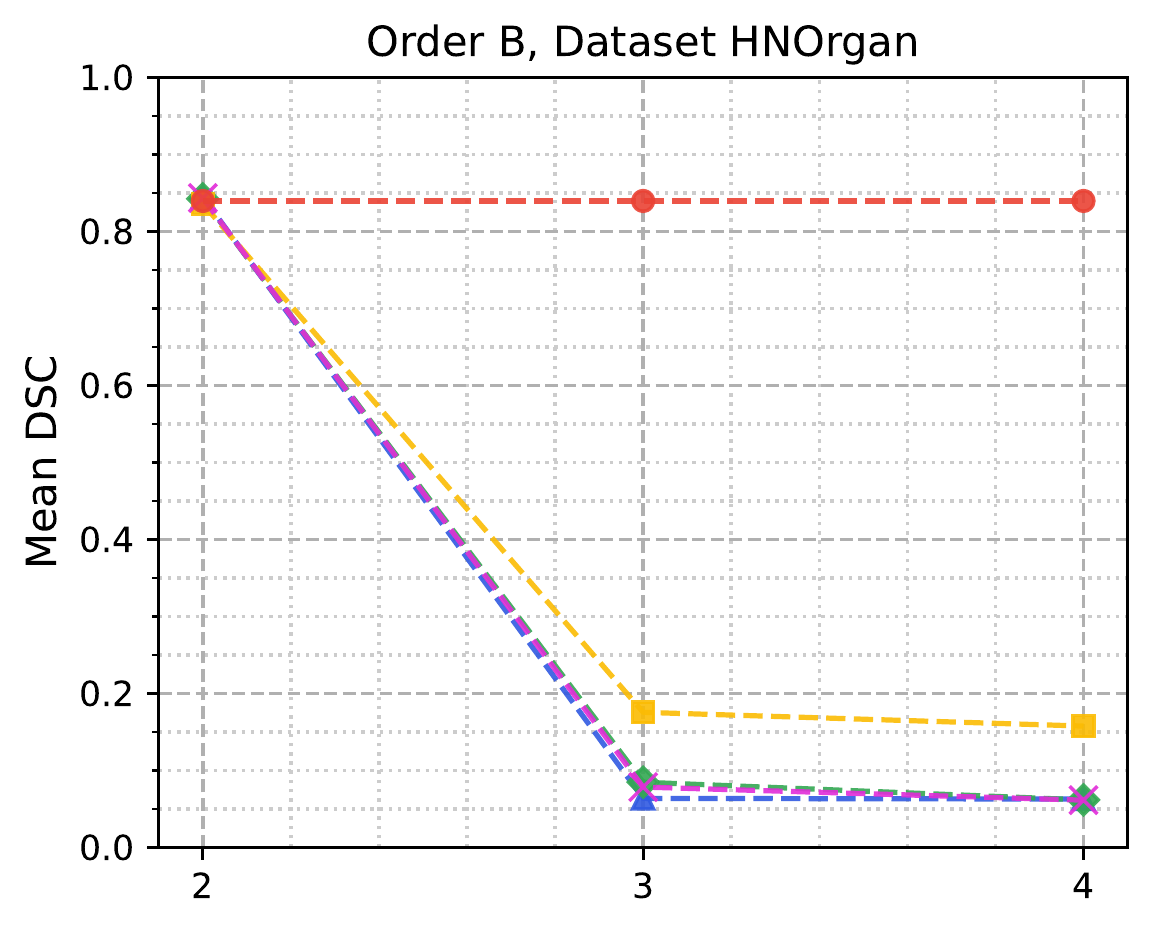}
    \caption{The mean DSCs over each dataset at each step of {\bf Order A} (solid line) and {\bf Order B} (dashed line).}\label{fig:step_dataset}
    \vspace{-3mm}
\end{figure*}

In our main text, we have briefly demonstrated and discussed the final segmentation performance and forgetting curve of our method and other comparison methods at each continual learning step in Table \ref{M-tab:main} and Fig \ref{M-fig:step_all} (Section 4.3). Here, we show the detailed numeric results of Fig \ref{M-fig:step_all} in Fig \ref{fig:step_dataset} and Table \ref{tab:step_task} and provide more in-depth discussion on results achieved by our model and other comparison methods. 

{\bf First}, the mean DSC and the forgetting rate in each dataset/step of two CSS orders are detailed in Table \ref{tab:step_task}. The mean DSC and average forgetting rate over all 143 organs at the last step is also shown. Besides the observations discussed in the main text, several additional findings can be noticed. First, the slight decrease of DSC of our method in the sub-figures of Fig \ref{M-fig:step_all} (mean DSC over all learned organs at the current step) is not due to our model forgetting, but simply because the achieved DSC values are lower in new datasets/steps. E.g., the upper bound mean DSC in ChestOrgan dataset is only 78.45\% (as compared to 93.24\% in TotalSegmentator dataset). As shown the last three rows of two CSS order in Table \ref{tab:step_task}, our method completely avoids forgetting of old knowledge when continually learning new dataset/task because of our proposed framework (frozen general encoder, light-weighted decoders, and body-part and anomaly-aware merging). In contrast, other distillation-based CSS methods all experienced severe forgetting with more than 50\% forgetting rate at the last step.

{\bf Second}, regarding the step-wise results of comparison methods. It is observed that methods based on the output-level knowledge distillation and MiB losses (MiB~\cite{cermelli2020modeling}, ILT~\cite{michieli2019incremental} and LISMO~\cite{liu2022learning}) suffer from catastrophic forgetting after the last step (overall mean DSC $<25\%$ and forgetting rate $>64\%$). In contrast, although PLOP~\cite{douillard2021plop} also has a large forgetting rate (about 50\%), the overall performance is noticeably better as compared to the other three methods. For instance, the overall mean DSC is 39.01\% for PLOP in CSS order A, which is at least 3 times higher than that for MiB, ILT and LISMO. The increased ability to keep old knowledge in PLOP might come from the applied entropy-based pseudo-labeling and the knowledge distillation on intermediate features in both encoder and decoder. 

{\bf Third}, regarding two continual segmentation orders, the main difference is the learning order of ChestOrgan and HNOrgan: order A first learns ChestOrgan in step-2, then HNOrgan in step-3, while order B exchanges the dataset in step-2 and 3. It is observed in Table~\ref{tab:step_task} and Fig \ref{fig:step_dataset} that the comparison methods forgetting rate at learning step-2 for the TotalSegmentator dataset is higher in order B than that in order A. E.g., TotalSegmentator DSC of MiB~\cite{cermelli2020modeling} at step-2 is decreased from 93.24\% $\rightarrow$ 21.96\% in order B vs. from 93.24\% $\rightarrow$ 45.80\% in order A. Notice that order B learns the HNOrgan right after TotalSegmentator at step-2, and HNOrgan contains CT images only focusing in the head and neck region, where TotalSegmentator has organs mostly labeled in the chest, abdomen, and pelvic regions. As a result, MiB can no longer see the chest, abdomen, and pelvic regions at step-2, which causes catastrophic forgetting in these body parts resulting in a significant performance drop. Instead, order A learns the ChestOrgan at step-2, and ChestOrgan covers all the chest and neck regions as well as most parts of the abdomen. Hence, MiB is still able to rehearse some old knowledge over these overlapping body parts so that the forgetting rate is reduced as compared to that in order B. A similar trend can be found in the forgetting curves of ChestOrgan and HNOrgan. These findings show that for the multi-organ continual segmentation, the forgetting rate of other comparison methods is closely related to the overlapping range of body parts in each dataset/step. In contrast, our proposed architectural-based method is learning-order and body-part invariant, which facilitates the model deployment in clinical practice.


{\bf Last}, we evaluate the impact of Alternative General Encoders. We recommended starting with TotalSegentator as it covers most body parts for comprehensive feature extraction. Alternatively, other datasets can be used as the starting dataset to train General Encoder. We trained the General Encoder using the ChestOrgan dataset. A tolerable performance drop ($<$1\% Dice) is observed in the CSS \textbf{Order A} final results. The assumed reason is that the torso region includes diverse anatomies and covers most of the body parts, and hence exhibits similar performance as the one trained using the TotalSegmentator dataset. Yet, when using the HNOrgan dataset to train the General Encoder, we notice a markedly $3\%$ Dice drop in the final results.

\section{Detailed Results of Individual Organs}
\label{supp:organ}
We provide detailed organ segmentation results of our final model as well as the upper bound nnUNet performance trained and evaluated on TotalSegmentator, ChestOrgan, HNOrgan datasets (shown in Table \ref{tab:totalseg_organ}, \ref{tab:chest_organ}, \ref{tab:hn_organ}). The final performance on the EsoOrgan dataet has been reported in the Table 1 of the main text. For organs in TotalSegmentator, due to the large amount of organs, we choose to group the 103 whole body organs into eight anatomical groups (Table \ref{tab:totalseg_group}) and report the average scores of each group (see Table \ref{tab:totalseg_organ}).
As shown in those tables, our final model performs closely to the upper bound accuracy when training separate nnUNet models on each dataset. There are no organs experiencing a large performance drop. The overall slightly drop in DSC and increasing in HD95/ASD is because of the decoder pruning process. 

\begin{table}
\centering
\caption{TotalSegmentator~\cite{wasserthal2022totalsegmentator} label list of each organ group.}
\label{tab:totalseg_group}
\vspace{1em}
\resizebox{0.8\columnwidth}{!}{%
\begin{tabular}{@{}llll@{}}
\toprule
\textbf{TotalSeg Organ Group}   & \multicolumn{3}{l}{\textbf{TotalSeg Organ Labels}} \\ \midrule
\textbf{Main Chest Organs}       & \multicolumn{3}{l}{13---17, 42---48}               \\
\textbf{Cardiovascular Vessels} & \multicolumn{3}{l}{7, 8, 9, 49, 51, 52, 53, 54}    \\
\textbf{Excretory Organs}       & \multicolumn{3}{l}{2, 3, 55, 57, 104}              \\
\textbf{Main Abdomen Organs}     & \multicolumn{3}{l}{1, 4, 5, 6, 10, 11, 12, 56}     \\
\textbf{Head}                   & \multicolumn{3}{l}{50 (`face'-93 is exluded)}      \\
\textbf{Vertebraes}             & \multicolumn{3}{l}{18---41}                        \\
\textbf{Ribs}                   & \multicolumn{3}{l}{58---81}                        \\
\textbf{Other Bones}            & \multicolumn{3}{l}{82---92}                        \\
\textbf{Muscles}                & \multicolumn{3}{l}{94---103}                       \\ \bottomrule
\end{tabular}%
}
\end{table}
\begin{table}
\caption{Mean DSC (\%, $\uparrow$), HD95 (mm, $\downarrow$) and ASD (mm, $\downarrow$) of 8 anatomical organ groups in TotalSeg (total 103 full body organs) of upper bound model and our final model.}
\label{tab:totalseg_organ}
\resizebox{\columnwidth}{!}{%
\begin{tabular}{@{}l|rrr|rrr@{}}
\toprule
\multicolumn{1}{c|}{\multirow{2}{*}{\textbf{TotalSeg Organ Group}}} & \multicolumn{3}{c|}{\textbf{Upper Bound}}                                                                & \multicolumn{3}{c}{\textbf{Ours}}                                                                       \\ \cmidrule(l){2-7} 
\multicolumn{1}{c|}{}                                               & \multicolumn{1}{c}{\textbf{DSC}} & \multicolumn{1}{c}{\textbf{HD95}} & \multicolumn{1}{c|}{\textbf{ASD}} & \multicolumn{1}{c}{\textbf{DSC}} & \multicolumn{1}{c}{\textbf{HD95}} & \multicolumn{1}{c}{\textbf{ASD}} \\ \midrule
\textbf{Chest Main Organ}                                           & 96.66                            & 1.71                              & 0.35                              & 95.62                            & 2.54                              & 0.43                             \\
\textbf{Cardiovascular Vessels}                                     & 91.75                            & 2.33                              & 0.55                              & 91.97                            & 2.96                              & 0.74                             \\
\textbf{Excretory System}                                           & 93.28                            & 4.24                              & 1.22                              & 93.53                            & 4.45                              & 1.28                             \\
\textbf{Abdomen Main Organ}                                         & 89.44                            & 3.45                              & 0.80                              & 91.05                            & 3.96                              & 0.88                             \\
\textbf{Head}                                                       & 94.51                            & 2.44                              & 0.62                              & 94.61                            & 3.02                              & 0.69                             \\
\textbf{Vertebraes}                                                 & 92.94                            & 2.01                              & 0.48                              & 92.65                            & 2.92                              & 0.68                             \\
\textbf{Ribs}                                                       & 91.54                            & 4.24                              & 1.06                              & 91.49                            & 5.15                              & 1.20                             \\
\textbf{Other Bones}                                                & 95.02                            & 6.94                              & 2.08                              & 93.14                            & 7.85                              & 2.27                             \\
\textbf{Muscles}                                                    & 96.09                            & 2.17                              & 0.38                              & 95.92                            & 2.98                              & 0.54                             \\ \bottomrule
\end{tabular}%
}
\end{table}


\begin{table*}
\centering
\caption{Mean and standard deviation of DSC (\%, $\uparrow$), HD95 (mm, $\downarrow$) and ASD (mm, $\downarrow$) of each ChestOrgan organ (total 31 chest organs) of upper bound model and our final model. \_L and \_R refer to the left and right.}
\label{tab:chest_organ}
\vspace{1em}
\scalebox{0.83}{
\begin{tabular}{@{}l|lll|lll@{}}
\toprule
\multirow{2}{*}{\textbf{ChestOrgan}} & \multicolumn{3}{c|}{\textbf{Upper Bound}}          & \multicolumn{3}{c}{\textbf{Ours}}                  \\ \cmidrule(l){2-7} 
                                     & \textbf{DSC}    & \textbf{HD95}   & \textbf{ASD}   & \textbf{DSC}    & \textbf{HD95}   & \textbf{ASD}   \\ \midrule
\textbf{Esophagus}                   & 85.41$\pm$4.12  & 4.93$\pm$3.32   & 0.68$\pm$0.42  & 85.22$\pm$4.30  & 5.78$\pm$2.81   & 0.67$\pm$0.34  \\
\textbf{Sternum}                     & 90.30$\pm$3.27  & 5.07$\pm$5.00   & 1.16$\pm$1.07  & 90.26$\pm$3.64  & 5.83$\pm$5.33   & 1.18$\pm$2.07  \\
\textbf{Thyroid\_L}                  & 84.13$\pm$4.71  & 2.87$\pm$1.36   & 0.66$\pm$0.36  & 84.10$\pm$3.98  & 3.54$\pm$1.23   & 0.73$\pm$0.33  \\
\textbf{Thyroid\_R}                  & 82.79$\pm$6.28  & 3.39$\pm$2.34   & 0.79$\pm$0.41  & 82.75$\pm$5.32  & 4.18$\pm$1.89   & 0.88$\pm$0.38  \\
\textbf{Trachea}                     & 93.74$\pm$2.19  & 4.74$\pm$3.45   & 1.09$\pm$0.88  & 93.68$\pm$1.83  & 5.45$\pm$2.76   & 1.03$\pm$0.69  \\
\textbf{Bronchus\_L}                 & 86.53$\pm$4.39  & 4.84$\pm$2.55   & 0.48$\pm$0.33  & 86.34$\pm$5.15  & 5.45$\pm$2.29   & 0.44$\pm$0.29  \\
\textbf{Bronchus\_R}                 & 75.86$\pm$13.79 & 9.18$\pm$7.51   & 2.35$\pm$2.87  & 75.88$\pm$13.41 & 10.42$\pm$7.67  & 2.41$\pm$2.70  \\
\textbf{Anterior cervical muscle}    & 69.23$\pm$8.05  & 6.31$\pm$5.67   & 1.34$\pm$1.22  & 69.02$\pm$6.79  & 7.08$\pm$4.54   & 1.39$\pm$1.20  \\
\textbf{Scalenus muscle}             & 74.26$\pm$4.42  & 5.83$\pm$3.79   & 0.83$\pm$0.38  & 74.24$\pm$5.18  & 6.98$\pm$3.86   & 0.79$\pm$0.38  \\
\textbf{Scalenus anterior muscle}    & 77.89$\pm$6.34  & 4.37$\pm$3.46   & 0.95$\pm$1.03  & 77.82$\pm$6.89  & 5.04$\pm$3.17   & 0.83$\pm$0.83  \\
\textbf{sternocleidomastoid muscle}  & 82.17$\pm$3.92  & 4.01$\pm$2.89   & 1.16$\pm$1.11  & 82.10$\pm$3.35  & 4.89$\pm$3.23   & 1.06$\pm$0.81  \\
\textbf{common carotid artery\_L}    & 78.09$\pm$7.65  & 9.33$\pm$17.74  & 2.17$\pm$4.33  & 77.97$\pm$8.21  & 9.98$\pm$16.47  & 2.28$\pm$4.33  \\
\textbf{common carotid artery\_R}    & 73.92$\pm$11.08 & 12.43$\pm$16.29 & 2.84$\pm$4.72  & 73.77$\pm$10.63 & 13.37$\pm$17.83 & 2.90$\pm$4.81  \\
\textbf{Pulmonary artery}            & 90.11$\pm$3.64  & 6.64$\pm$3.09   & 1.28$\pm$0.67  & 89.93$\pm$5.46  & 7.11$\pm$3.97   & 1.13$\pm$0.77  \\
\textbf{Subclavian artery\_L}        & 71.79$\pm$9.26  & 20.26$\pm$20.29 & 2.37$\pm$3.40  & 71.78$\pm$9.50  & 23.15$\pm$22.32 & 2.28$\pm$3.46  \\
\textbf{Subclavian artery\_R}        & 80.03$\pm$5.87  & 11.93$\pm$13.24 & 2.04$\pm$2.11  & 79.97$\pm$6.07  & 12.14$\pm$13.72 & 2.19$\pm$2.90  \\
\textbf{Vertebral artery\_L}         & 47.72$\pm$19.66 & 20.25$\pm$21.22 & 6.65$\pm$11.83 & 46.62$\pm$20.32 & 21.27$\pm$20.94 & 6.51$\pm$11.73 \\
\textbf{Vertebral artery\_R}         & 45.28$\pm$18.38 & 19.90$\pm$21.07 & 6.33$\pm$11.35 & 42.66$\pm$17.08 & 20.81$\pm$17.43 & 6.40$\pm$11.12 \\
\textbf{Ascending aorta}             & 93.08$\pm$2.54  & 5.41$\pm$2.40   & 1.39$\pm$0.77  & 93.05$\pm$2.17  & 6.25$\pm$2.50   & 1.34$\pm$0.97  \\
\textbf{Descending aorta}            & 97.13$\pm$1.75  & 3.21$\pm$2.21   & 0.72$\pm$0.37  & 97.05$\pm$0.94  & 3.73$\pm$1.98   & 0.67$\pm$0.35  \\
\textbf{Aorta arch}                  & 92.12$\pm$9.08  & 4.23$\pm$2.72   & 1.25$\pm$1.42  & 92.11$\pm$7.53  & 4.93$\pm$2.80   & 1.27$\pm$1.42  \\
\textbf{Azygos vein}                 & 73.29$\pm$11.53 & 8.99$\pm$14.07  & 1.68$\pm$3.58  & 73.25$\pm$10.02 & 9.79$\pm$14.16  & 1.65$\pm$3.53  \\
\textbf{brachiocephalic vein\_L}     & 85.80$\pm$5.57  & 3.27$\pm$2.35   & 0.35$\pm$0.29  & 85.73$\pm$5.00  & 4.13$\pm$2.76   & 0.35$\pm$0.26  \\
\textbf{brachiocephalic vein\_R}     & 85.83$\pm$5.07  & 4.71$\pm$1.88   & 0.87$\pm$0.54  & 85.82$\pm$5.70  & 5.30$\pm$2.37   & 0.89$\pm$0.57  \\
\textbf{internal jugular vein\_L}    & 74.66$\pm$14.57 & 12.77$\pm$16.31 & 2.88$\pm$4.51  & 74.63$\pm$14.41 & 14.19$\pm$14.63 & 2.90$\pm$3.81  \\
\textbf{internal jugular vein\_R}    & 78.28$\pm$8.73  & 12.03$\pm$16.86 & 2.95$\pm$3.68  & 78.24$\pm$8.19  & 15.86$\pm$13.63 & 3.02$\pm$3.52  \\
\textbf{Pulmonary vein}              & 70.62$\pm$8.24  & 7.04$\pm$2.80   & 1.53$\pm$0.67  & 70.57$\pm$7.83  & 7.81$\pm$2.91   & 1.49$\pm$0.76  \\
\textbf{Subclavian vein\_L}          & 63.32$\pm$14.72 & 9.64$\pm$6.69   & 1.99$\pm$1.90  & 63.32$\pm$15.21 & 10.14$\pm$7.88  & 2.00$\pm$1.71  \\
\textbf{Subclavian vein\_R}          & 60.59$\pm$12.21 & 11.62$\pm$8.13  & 2.75$\pm$2.35  & 60.60$\pm$12.92 & 13.89$\pm$10.63 & 2.83$\pm$2.20  \\ 
\textbf{Inferior vena cava}          & 82.51$\pm$6.23  & 7.95$\pm$5.58   & 1.69$\pm$1.13  & 82.45$\pm$5.82  & 9.01$\pm$5.56   & 1.71$\pm$1.25  \\
\textbf{Superior vena cava}          & 85.38$\pm$3.83  & 5.85$\pm$3.55   & 1.36$\pm$0.88  & 85.29$\pm$3.85  & 6.60$\pm$3.45   & 1.33$\pm$0.75  \\
\bottomrule
\end{tabular}%
}
\end{table*}

\begin{table*}
\centering
\caption{Mean and standard deviation of DSC (\%, $\uparrow$), HD95 (mm, $\downarrow$) and ASD (mm, $\downarrow$) of each HNOrgan organ (total 13 head-neck organs) of upper bound model and our final model.  \_L and \_R refer to the left and right.}
\label{tab:hn_organ}
\vspace{1em}
\scalebox{0.83}{
\begin{tabular}{@{}l|lll|lll@{}}
\toprule
\multirow{2}{*}{\textbf{HNOrgan}} & \multicolumn{3}{c|}{\textbf{Upper Bound}}       & \multicolumn{3}{c}{\textbf{Ours}}               \\ \cmidrule(l){2-7} 
                                  & \textbf{DSC}    & \textbf{HD95} & \textbf{ASD}  & \textbf{DSC}    & \textbf{HD95} & \textbf{ASD}  \\ \midrule
\textbf{BrainStem}                & 91.42$\pm$2.47  & 2.93$\pm$1.15 & 0.76$\pm$0.32 & 90.89$\pm$2.38  & 2.65$\pm$1.37 & 0.74$\pm$0.36 \\
\textbf{Eye\_L}                   & 92.32$\pm$1.73  & 1.57$\pm$0.60 & 0.36$\pm$0.11 & 92.25$\pm$1.69  & 1.50$\pm$0.48 & 0.33$\pm$0.12 \\
\textbf{Eye\_R}                   & 91.99$\pm$1.49  & 1.68$\pm$0.54 & 0.40$\pm$0.10 & 91.95$\pm$1.19  & 1.63$\pm$0.92 & 0.49$\pm$0.21 \\
\textbf{Lens\_L}                  & 81.49$\pm$10.64 & 1.64$\pm$0.90 & 0.53$\pm$0.47 & 80.33$\pm$8.81  & 1.55$\pm$0.89 & 0.47$\pm$0.50 \\
\textbf{Lens\_R}                  & 84.16$\pm$8.35  & 1.39$\pm$0.70 & 0.40$\pm$0.30 & 82.46$\pm$6.33  & 1.23$\pm$1.12 & 0.36$\pm$0.36 \\
\textbf{Optic Chiasm}             & 67.04$\pm$13.34 & 3.73$\pm$1.66 & 1.03$\pm$0.65 & 66.60$\pm$13.29 & 3.59$\pm$1.91 & 0.96$\pm$0.71 \\
\textbf{Optic Nerve\_L}           & 74.34$\pm$6.77  & 3.05$\pm$2.75 & 0.56$\pm$0.27 & 74.72$\pm$6.93  & 2.96$\pm$3.27 & 0.61$\pm$0.51 \\
\textbf{Optic Nerve\_R}           & 75.15$\pm$6.47  & 2.64$\pm$0.80 & 0.51$\pm$0.30 & 73.64$\pm$7.07  & 2.49$\pm$1.18 & 0.56$\pm$0.35 \\
\textbf{Parotid\_L}               & 91.32$\pm$3.01  & 3.13$\pm$1.72 & 0.75$\pm$0.33 & 91.09$\pm$3.08  & 2.90$\pm$2.10 & 0.76$\pm$0.32 \\
\textbf{Parotid\_R}               & 90.93$\pm$2.87  & 3.22$\pm$1.94 & 0.86$\pm$0.46 & 90.94$\pm$2.97  & 3.03$\pm$2.10 & 0.86$\pm$0.67 \\
\textbf{TMJ\_L}                   & 81.55$\pm$9.42  & 2.10$\pm$1.06 & 0.56$\pm$0.47 & 82.14$\pm$9.33  & 1.87$\pm$1.19 & 0.52$\pm$0.65 \\
\textbf{TMJ\_R}                   & 84.81$\pm$8.57  & 1.85$\pm$1.06 & 0.43$\pm$0.39 & 84.70$\pm$9.28  & 1.69$\pm$0.93 & 0.42$\pm$0.50 \\
\textbf{Spinal cord}              & 90.01$\pm$2.35  & 2.03$\pm$0.61 & 0.65$\pm$0.22 & 89.93$\pm$2.19  & 1.83$\pm$0.81 & 0.64$\pm$0.31 \\ \bottomrule
\end{tabular}%
}
\end{table*}

\section{Re-implementation of Comparison Methods}
\label{supp:impl_comp}
For all comparison methods, we start with the same pretrained nnUNet model on TotalSegmentator dataset, which has been trained using 3D nnUNet setting for 8000 epochs, with 250 iterations per epoch and initial learning rate 0.01. After that, the model is finetuned sequentially on continual segmentation tasks (ChestOrgan, HNOrgan and EsoOrgan), where each dataset are finetuned for 500 epochs, with 250 iterations per epoch and initial learning rate 0.005. All the other nnUNet settings, such as data augmentation, remain the same as our implementation. Moverover, since our segmentation datasets/tasks are 3D CT scans (different from the previous continual segmentation works in natural images), adjustments to these comparison methods are required (extending 2D methods to 3D), as well as transferring their implementations to the nnUNet framework. We describe the detailed re-implementation of previous methods, especially our modifications, in the following subsections. 


\noindent\textbf{MiB.}
MiB~\cite{cermelli2020modeling} proposes two marginal losses, or unbiased losses to solve the background shift issue in continual segmentation in their original paper: unbiased cross-entropy (UNCE) loss, which merges the probabilities of old classes to the background label, and unbiased knowledge distillation (UNKD) loss, which merges the probabilities of all new classes (belonging unseen classes of the old model) to the background label. Notice that, the original unbiased loss assumes that new classes from the current dataset are completely disjoint with all the old classes, however, this assumption is not holding in our datasets. E.g., TotalSegmentator and ChestOrgan contain four overlapping organs: inferior vena cava, trachea, esophagus and pulmonary artery. Therefore, in order to re-implement MiB losses in the nnUNet framework and make them compatibility with our datasets, we slightly modifies and generalizes both unbiased losses to handle overlapping labels in the continual learning setting. The modified UNCE loss is as follows: 

\begin{equation}
    \ell^{\theta^t}_{ce}(x,y) = -\frac{1}{\lvert \mathcal{I} \rvert}\sum_{i\in \mathcal{I}}\log \tilde{q}_x^t (i,y_i)
\end{equation}

\noindent where:

\begin{equation}\label{eq:unce}
    \tilde{q}_x^t (i,c) = 
    \begin{cases}
        q_x^t(i,c) &\text{if $c \neq b$}\\
        \sum_{k\in \gY^{t-1} \textcolor{red}{\setminus \gC^t_{p}}} q_x^t(i,k) &\text{if $c = b$}
    \end{cases}
\end{equation}

\noindent Here, same notations referred to the original paper is used, except $\gC^t_{p}=\gY^{t-1} \cup \gC^t - b$, which indicates the overlapping classes (excluding background label) between current dataset $\gC^t$ and all the previous classes $\gY^{t-1}$ at the learning step $t$. When calculating UNCE loss, we merge all the old labels to the background except the overlapping classes. 

Similarly, we adapt UNKD loss as: 

\begin{equation}
    \ell^{\theta^t}_{kd}(x,y) = -\frac{1}{\lvert \mathcal{I} \rvert}\sum_{i\in \mathcal{I}} \sum_{c \in \gY^{t-1} \textcolor{red}{\setminus \gC^t_p}} q_x^{t-1}(i,c) \log \hat{q}_x^t (i,y_i)
\end{equation}

\noindent where:

\begin{equation}\label{eq:unkd}
    \hat{q}_x^t (i,c) = 
    \begin{cases}
        q_x^t(i,c) &\text{if $c \neq b$}\\
        \sum_{k\in \gC^t} q_x^t(i,k) &\text{if $c = b$}
    \end{cases}
\end{equation}

\noindent In the above formula, overlapping organs from the old class set are excluded so that the knowledge distillation works on the real old classes that cannot be learned from the current dataset. 

Using two modified losses, we always train the model with the latest labels and ignore the previously learned overlapping labels when overlapping organs occur. Thus, overlapping labels are trained directly using the cross-entropy loss and merged to the background in the knowledge distillation loss. In addition, we use the same hyperparameters as the MiB setting: the weight of UNKD loss are set as 10 with balanced classifier initialization strategy.

\noindent\textbf{ILT.}
ILT~\cite{michieli2019incremental} originally first proposes the continual semantic segmentation (CSS) protocol and provides a naive solution using an output-level knowledge distillation on the old classes ($\gL_D'$) and a feature-level knowledge distillation on the intermediate features from encoder ($\gL_D''$). This method leads to inferior performance and experienced severe forgetting as compared to MiB and other CSS methods on multiple natural image benchmarks~\cite{douillard2021plop,michieli2021continual}. 
In order to improve ILT performance on our datasets/tasks, we modifies the original ILT setting and losses as follows: (1) ILT uses a frozen encoder setting ($E_F$) together with $\gL_D'$, which is similar to our general encoder method, therefore, we re-implement ILT using this frozen encoder setting, as mentioned in the main paper; (2) since original ILT losses do not alleviate the background shift issue and have a large bias towards new classes (experiencing severe forgetting even with the frozen encoder), we additionally apply the MiB loss (Eq.\ref{eq:unce},\ref{eq:unkd}) to reinforce the decoder to preserve more old knowledge. In short, our re-implemented ILT can be treated as a frozen encoder version of MiB ($E_F + \gL_{\text{mib}}$). Although leading to an improved performance as compared to the original ILT, this frozen encoder ILT version still has obvious knowledge forgetting as shown in Table 1 of the main text. This indicates that the frozen encoder with unbiased output-level knowledge distillation is not sufficient to preserve the old knowledge in CSS. In contrast, our proposed framework (general encoder + light-weighted decoder) can performance at the accuracy for the first time with real non-forgetting in CSS. 


\noindent\textbf{PLOP.}
PLOP~\cite{douillard2021plop} is originally implemented for 2D images, especially its multi-scale local distillation loss based on local POD. Local POD is a multi-scale feature pooling strategy consisting of computing width and height-pooled slices on multi-scale regions, which aims to better retain both global and local spatial knowledge from the old model. However, since our data are all 3D CT scans with an additional depth dimension, we specifically extend the local POD to higher dimensions when re-implementing the method.  Two pooling strategies can be adopted for the 3D cases: (1) pooling 3D feature map along each single dimension and extracting three 2D projections along each axis: 

\begin{align}
    \begin{split}
    \Phi(\rvx)=
    &\left( \frac{1}{H}\sum_{h=1}^{H} \rvx\left[ h,:,:,: \right] \middle\| \frac{1}{W}\sum_{w=1}^{W} \rvx\left[ :,w,:,: \right] \right. \\ 
    &\left. \middle\| \frac{1}{D}\sum_{d=1}^{D} \rvx\left[ :,:,d,: \right] \right) \in \gR^{(WD + HD + HW) \times C}
    \end{split}
\end{align}

\noindent where notations follow the original PLOP paper. This pooling method can preserve enough spatial information meanwhile providing some level of plasticity to the model. (2) Pooling 3D feature map on two dimensions and only extract 1D projection along the remaining axis: 

\begin{align}
    \begin{split}
    \Phi(\rvx)=
    &\left( \frac{1}{HW}\sum_{h=1}^{H}\sum_{w=1}^{W} \rvx\left[ h,w,:,: \right] \middle\| \frac{1}{WD}\sum_{w=1}^{W}\sum_{d=1}^{D} \rvx\left[ :,w,d,: \right] \right. \\ 
    &\left. \middle\| \frac{1}{HD}\sum_{h=1}^{H}\sum_{d=1}^{D} \rvx\left[ h,:,d,: \right] \right) \in \gR^{(H+W+D) \times C}
    \end{split}
\end{align}

\noindent This pooling strategy has similar feature shape, however, when pooling on two axes together, most of the spatial information are lost and POD loss cannot retain the old knowledge. After comparing the performance using two strategies, we select the former one, which better handles the trade-off between model rigidity and plasticity. 

For hyperparameters, the original paper uses the pod weighting factor of 0.01, which is too large for the 3D pooling case. Because the L2 norm of 3D pooled features is more than 10 times larger than that of 2D pooled features. In our experiments, we set this pod factor to 0.001. Other hyperparameters are consistent with those used in the original paper.

\noindent\textbf{LISMO.}
The original LISMO~\cite{liu2022learning} is designed based on nnUNet framework, so we are able to directly re-implement this method. We would like to mention several important differences between our datasets and those used in LISMO. In LISMO~\cite{liu2022learning}, it has a slightly improved result than MiB when segmenting five large abdominal organs in their experiment (using 3D low resolution of nnUNet). Under this setup, all five abdominal organs could be seen in each 3D training patch most of the time, which could frequently reinforce and rehearsal the model's ability on unseen organs in the current dataset through their memory module and prototype matching loss. However, this is not the case in our experiments, since many old organs are no longer able to observe in the new dataset due to non-overlapping body parts. E.g. abdominal organs cannot appear in the HNOrgan dataset. Moreover, the high resolution nnUNet version is used to meet the high segmentation accuracy required in practice and there are over 100 target organs spreading among the whole body range, so our patch size is impossible to cover most organs within each patch. Under this situation, the prototype matching loss is not able to compute on non-existing organs and the contrastive loss is not sufficient to constraint the feature distributions of these organs, which results in severe forgetting for the unobserved organs in our experiment.



\end{document}